\documentclass[runningheads]{llncs}

\usepackage{eccv}

\usepackage{booktabs} 
\usepackage{multirow} 
\usepackage{adjustbox}
\usepackage{tikz}
\usetikzlibrary{calc}
\usepackage{svg}
\usepackage{enumitem}
\usepackage{float}
\usepackage[hidelinks]{hyperref}

\usepackage{eccvabbrv}

\usepackage{graphicx}
\usepackage{booktabs}
\usepackage{xcolor}

\usepackage[accsupp]{axessibility}  

\usepackage{hyperref}

\usepackage{orcidlink}

\begin{document}

\title{Triangle Splatting SLAM} 

\titlerunning{Triangle Splatting SLAM}

\author{Nicholas Fry\inst{1, 2} \and
Eric Dexheimer\inst{2} \and
Kirill Mazur\inst{2} \and 
Paul H.~J.~Kelly\inst{1, 2}  \and 
Andrew J.~Davison\inst{2}}

\authorrunning{N.~Fry et al.}

\institute{Software Performance Optimisation Group, Imperial College London \and
Department of Computing, Imperial College London\\
\email{\{nf20, e.dexheimer21, k.mazur21, p.kelly, a.davison\}@imperial.ac.uk}}

\maketitle

\vspace{4mm}
Project page: \url{https://nmjfry.github.io/triangle-splatting-slam/}
\vspace{-7mm}

\begin{figure}[h!]
    \centering
    \includegraphics[width=\textwidth]{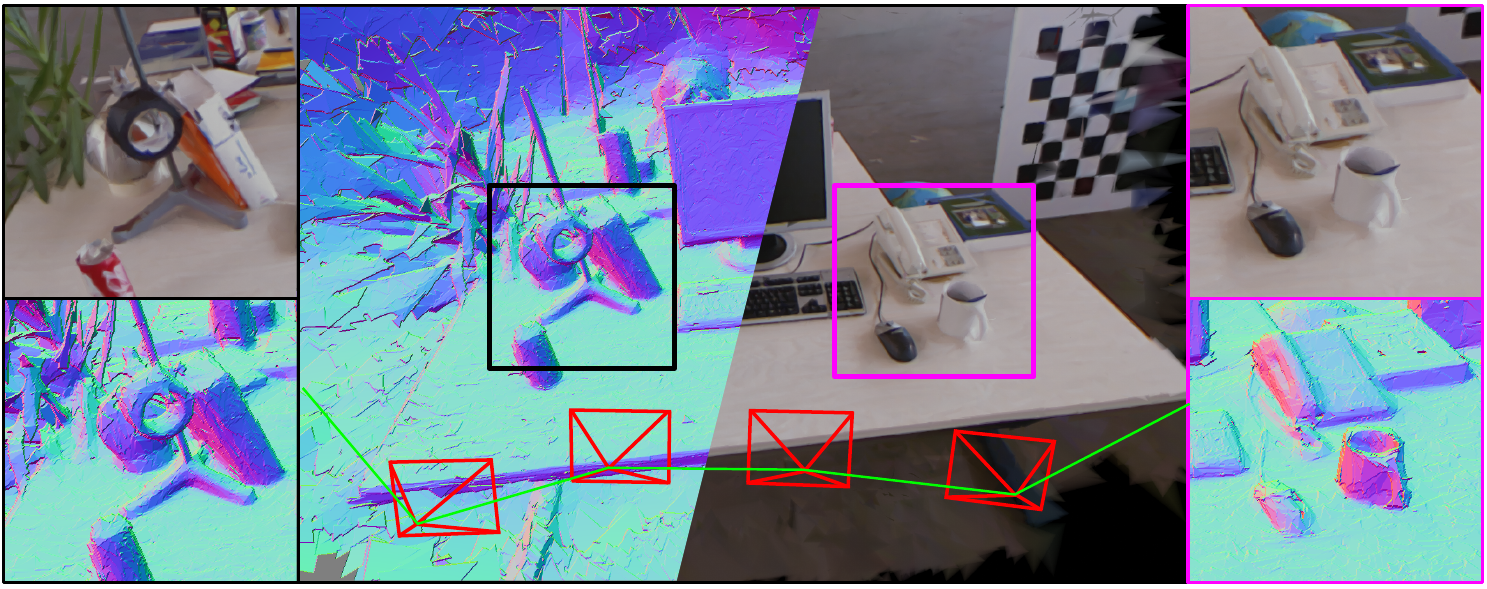} 
    \caption{Triangle Splatting SLAM uses triangles as the underlying scene representation to enable photo-realistic and high-fidelity geometry reconstruction, accurate camera pose estimation, and on-the-fly  mesh generation.}
    \label{fig:teaser}
\end{figure}

\begin{abstract} 

We present a dense RGB-D SLAM system using differentiable triangles as the 3D map representation. While 3D Gaussian Splatting has emerged as the leading method for novel-view synthesis, triangles remain the standard primitive for traditional rendering hardware, game engines, and downstream tasks requiring explicit geometry such as simulation, collision, and editing. Recent offline methods have demonstrated that an unstructured `triangle soup' can be optimised into a photorealistic mesh via Delaunay triangulation across a set of posed images. Building upon this insight, we present the first dense   SLAM system to employ Triangle Splatting to perform both tracking and mapping through online differentiable rendering of a triangle soup. The map can be converted into a connected mesh on-the-fly via restricted Delaunay triangulation, enabling new online capabilities such as mesh deformation and collision checking. On Replica and TUM-RGBD, our system outperforms baselines on 3D geometry, matches the camera-tracking accuracy, and enables online mesh-based scene editing. 

\end{abstract}

\section{Introduction}
\label{sec:intro}

Embodied AI applications such as robotics can benefit strongly from explicit 3D scene representations built incrementally as a camera moves through the world. 

Visual SLAM has evolved from sparse landmark maps to dense photorealistic reconstructions, with differentiable rendering enabling joint optimisation of camera poses and scene geometry. More recently, strong learned priors have dramatically improved depth and camera pose estimation. However, there is still much debate about the right representation for 3D in such systems, and none suggested so far offers all of the ideal properties we might want. Ideally, such a representation would be photorealistic, geometrically precise for simulation and planning, compact in memory, fast to render, straightforward to update incrementally, and flexible enough to accommodate scene changes; yet no existing approach achieves all of these goals simultaneously.

For decades, triangle meshes have been the industry standard for 3D scene representation in games, films, and other digital media, with GPUs optimised specifically to render them at high performance. Beyond rendering speed, meshes represent surface geometry explicitly and adaptively, with high precision in detailed areas and efficiency in simpler structures. Furthermore, their built-in connectivity makes them straightforward to edit, deform, and warp.

However, the connectivity that makes meshes so valuable for downstream tasks proves to be a significant liability when it comes to incremental reconstruction. Maintaining topological consistency while continuously integrating new sensor measurements is notoriously difficult, and for this reason dense visual SLAM methods have usually operated with other representations such as voxel or Signed Distance Function (SDF) grids, implicit neural functions, or disconnected clouds of point-like primitives such as surfels or 3D Gaussians. A triangle mesh is then generated if desired, but only as a secondary representation not directly used by the SLAM algorithm \cite{kinectfusion, schops2019surfelmeshing, schops_badslam_2019}.

In methods that use volumetric grids, meshing is achieved via the marching cubes algorithm \cite{marchingcubes}. The mesh must therefore be completely regenerated each time the map changes using a set of fixed 3D grid resolutions. Meshes produced indirectly from primitives such as 3D or 2D Gaussians \cite{kerbl20233d, matsuki2024gaussian, yu2024gaussian} require ray-marching or density-based SDF extraction, meaning the depth estimates don't necessarily correspond to the final mesh and thus does not guarantee geometric consistency. In 3D point-based triangulation methods, such as Delaunay triangulation, existing points are connected to form tetrahedra with triangular faces. Given a set of estimated 3D points derived from depth measurements or a prior, Delaunay triangulation maintains the existing geometry rather than generating a new set of vertices. Consequently, the method is inherently adaptive to the resolution of the input data, preserves the precise positions of these estimates, and supports incremental updates as opposed to full mesh regeneration.

Recent offline methods have shown that the 3DGS rendering pipeline can be adapted to triangle primitives \cite{Held2026MeshSplatting, tojo2026diffsoup}, representing a scene as a soup of triangles with learnable per-vertex colour and opacity rather than a connected mesh. Subsequent works show that one can then recover a photorealistic mesh using Delaunay triangulation of either this triangle soup, 3D Gaussian `pivots' or implicit features, avoiding the need for intermediate fusion, post-processing or deep pre-trained priors \cite{guedon2025milo, Held2026MeshSplatting, mai2026radiance}. 

In this paper, we present the first SLAM system to employ differentiable triangles \cite{Held2026MeshSplatting, held2025triangle} as its sole map representation, which can operate with and without interleaved meshing. Rather than deferring to a secondary implicit structure, our system models the scene directly as a set of vertices and faces. Both tracking and mapping are driven by differentiable rendering, where the rendered pixel colours are computed via interpolation of vertex features across the projected faces. By optimising the geometry directly, we achieve a photorealistic reconstruction that enables connected online mesh extraction through Delaunay triangulation.

In summary, our contributions are:
\begin{itemize}[label=\textbullet]
    \item A SLAM system using differentiable triangles as the underlying map representation.
    \item Novel strategies for achieving consistent online mapping from a stream of images, including camera pose optimisation, equilateral regularisation, and window-based densification.
    \item Competitive camera tracking quality on both Replica and TUM-RGBD datasets while achieving significantly higher 3D map quality over alternative differentiable rendering SLAM systems. 
    \item Online scene editing capabilities unlocked by the triangle mesh representation.
\end{itemize}

\section{Related Work}

\subsection{Representations for Differentiable Rendering}

The landscape of representations for 3D modeling is vast and the choice of representation fundamentally dictates the capabilities of the downstream task. While triangles have been dominant in the field of computer graphics, the inverse problem of reconstructing geometry from sensor readings has not reached consensus on the ideal representation. 

By making rendering differentiable, scene geometry and appearance can be optimised jointly from image observations alone.  NeRF-based representations \cite{mildenhall2020nerf, barron2022mip, mueller2022instant} are attractive for their high-fidelity view synthesis and  ability to compress information. However, their rendering speed is limited by expensive ray-marching and MLP queries which are prohibitive in online systems.  The implicit nature of these representations complicates geometric interpretation, editing, and integration with classical graphics and physics pipelines which hinders their adoption in intelligent embodied systems.

3D Gaussian Splatting (3DGS) \cite{kerbl20233d} uses anisotropic Gaussian primitives rendered via differentiable alpha compositing to achieve real-time rendering with exceptional visual fidelity. While splatting methods significantly improve rendering speed compared to neural implicit models, most remain fundamentally volumetric and do not directly yield structured surface representations. Consequently, their suitability for tasks requiring explicit geometry, such as physical simulation, lighting effects, collision detection, deformations or mesh-based editing, remains limited. Other methods, such as Gaussian Opacity Fields and 2D Gaussian Splatting \cite{yu2024gaussian, huang20242d} use ray-based intersection to better approximate surface geometry but require post-processing and are still not natively compatible with game engines.

Recent works demonstrate that high-quality meshes can be recovered using differentiable rasterisation on the Delaunay triangulation of structure-from-motion points \cite{guedon2025milo, Held2026MeshSplatting, mai2026radiance}, enabling offline photorealistic mesh reconstruction from posed images. Mesh-In-the-Loop (MILo) \cite{guedon2025milo} trains a set of 3D Gaussians before performing Delaunay triangulation on points sampled around the means of high-opacity Gaussians. A mesh is then extracted via Deep Marching Tetrahedra (DMTet) \cite{shen2021dmtet} on the Delaunay tetrahedra, and jointly supervised with the Gaussians to achieve a watertight, smooth, and photorealistic result. While the reconstruction quality is high, the process is slow, uses multiple representations, employs two triangulation strategies, and incurs significant memory cost, making it unsuitable for SLAM. Radiance Meshes \cite{mai2026radiance} similarly achieve high-quality extraction by storing an Instant-NGP \cite{mueller2022instant} within the Delaunay tetrahedra to model view-dependent appearance, but are also prohibitively slow to train. Mesh Splatting \cite{Held2026MeshSplatting} takes a more lightweight approach, optimising an alpha-blended triangle soup before applying Restricted Delaunay triangulation to recover a mesh with corrected connectivity. In a post-Delaunay training phase, vertices are further optimised for photorealism, which degrades topological consistency and reintroduces self-intersections. Although the final mesh quality is lower than other methods, the significantly reduced computational cost makes it a compelling foundation for online reconstruction. 
Radiant Foam \cite{govindarajan2025radiant} also performs 3D Delaunay tetrehedralisation, however it takes an entirely different approach to rendering; performing volumetric ray tracing through Voronoi cells to enable ray-based effects. Its sensitivity to initialisation and incompatibility with other rasterisation pipelines limit its applicability.

\subsection{Dense RGB-D SLAM}

Sparse RGB-D SLAM achieves accurate pose estimation in real-time \cite{mur-artal_orb-slam2_2017}. However, these methods lack a dense map which is required for augmented reality (AR) and robotics.  It is possible to treat mapping as a problem conditioned on already known poses.  For example, a voxel grid representing the truncated signed distance field (TSDF) integrates depth maps given the current pose estimate \cite{curless_tsdf_1996}.  This was first shown for RGB-D sensors in KinectFusion, which also uses the mesh from marching cubes for tracking \cite{kinectfusion}.  Storing a dense grid is limited by memory, resulting in a trade-off between efficiency and resolution.  
Kintinuous \cite{whelan_kintinuous_2013} addresses this by using a circular buffer for egocentric TSDF reconstruction, with portions exiting the grid converted into a mesh for large-scale reconstruction. However, decoupling the map into separate representations, one active and the other inactive, means previously visited areas cannot be re-used for tracking or mapping.

Furthermore, sparse voxel structures such as voxel hashing \cite{niessner_hashing_2013} enable scalability to larger areas, but early integration of measurements and the lack of joint optimisation of the poses and map can result in accumulated error leading to map corruption. Point-based representations emerged as an alternative to handle larger scenes in real-time while also continuously fusing new measurements into the current map \cite{pointbasedfusionKeller}.  To handle drift, ElasticFusion \cite{whelan_elasticfusion_2015} opts for a map-centric approach using surfels that is capable of deforming the entire map to correct for drift.  BAD-SLAM \cite{schops_badslam_2019} jointly optimises poses and surfels, which greatly improves upon the pose accuracy of dense methods that traditionally factorised the estimation.  DROID-SLAM achieves state-of-the-art robustness in both monocular and RGB-D scenarios by performing a learned dense bundle adjustment over depth maps.  However, unstructured points are not suitable for interaction and bundle adjustment does not optimise for coherent geometry. While surfels provide local surface information, the map is not suitable for physics simulation and true interaction.  SurfelMeshing \cite{schops2019surfelmeshing} avoids the uniformity of voxel grids and the re-integration of all depths into a volume during loop closure by asynchronously meshing on top of updates to a surfel map.  Ultimately, this still requires two representations, with the mesh not informing the geometry of the surfels themselves.

With the rise of differentiable rendering, forward models can be used to continuously update map parameters via back-propagation, which avoids the early integration of backward sensor models and the rigid updates imposed by previous data structures. Inspired by NeRF, iMAP \cite{sucar2021imap} optimises a coordinate MLP via rendered RGB and depth within a parallel tracking and mapping framework inspired by PTAM \cite{klein_parallel_2007}. NICE-SLAM \cite{zhu_niceslam_2022} improves scalability by replacing the global MLP with hierarchical voxel feature grids decoded by MLPs. Differentiable point-based representations \cite{sandstrom_point_2023} avoid the need to choose such a discretisation a priori, allocating capacity adaptively where the scene requires it rather than committing to a fixed grid resolution.

Gaussian Splatting SLAM \cite{matsuki2024gaussian, keetha_splatam_2024, yan2024gs} brought photorealistic real-time mapping to dense SLAM, but the volumetric nature of 3D Gaussians means geometry must be extracted indirectly via density fields, requiring post-processing and failing to guarantee accurate surfaces. Methods using 2D Gaussians with normal supervision \cite{matsuki20254dtam} move closer to an explicit surface representation, but like surfels, remain disconnected primitives that are not directly suitable for editing, simulation, or integration with standard graphics pipelines.

\section{Method}

\begin{figure}[t]
    \centering
    \includegraphics[width=\linewidth]{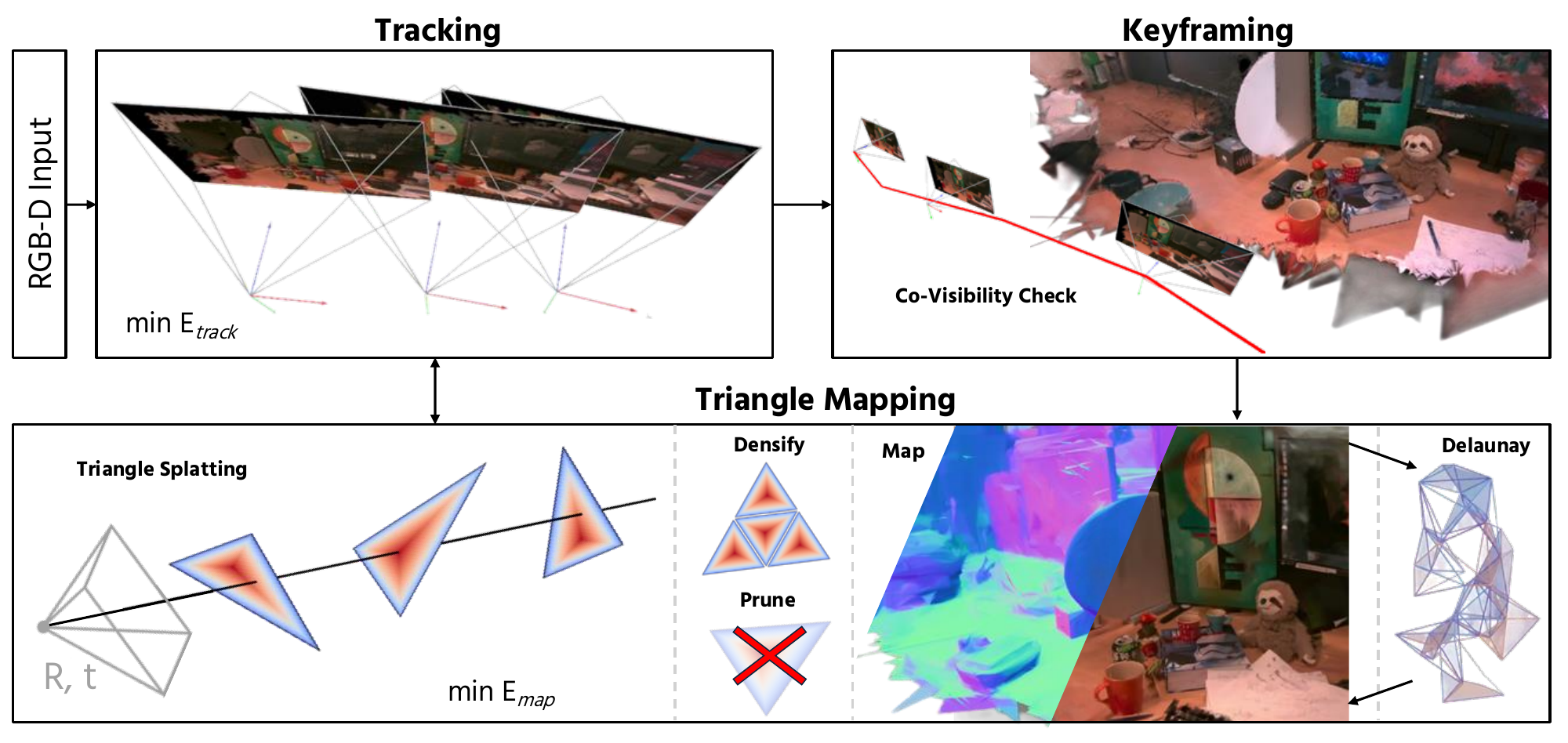}
    \caption{SLAM System Overview. \textbf{Tracking (top left)}: RGB-D frames are processed to estimate camera poses. \textbf{Keyframing (top right)}: Frames with high co-visibility are selected as keyframes. \textbf{Mapping (bottom)}: A triangle-based map is continually optimised via triangle splatting, with adaptive densification and pruning. Triangles can be converted into a connected mesh through restricted Delaunay triangulation.}
    \label{fig:systemarch}
\end{figure}

We provide an overview of our method, including the triangle rendering definition and its differentiable form, the addition of pose optimisation, and the full SLAM system using a tracking and mapping framework.  A system diagram is given in \cref{fig:systemarch}. 

\subsection{Mesh Splatting}

The map representation for our SLAM system builds upon Mesh Splatting \cite{Held2026MeshSplatting}, where the scene is represented by a set of $N$ vertices $\mathcal{V}$.

Each vertex $\mathbf{v}_i \in \mathcal{V}$, $i = 1, \dots, N$ is parameterised as $\mathbf{v}_i = (x_i, y_i, z_i, \mathbf{c}_i, \mathbf{o}_i)$; the 3D vertex position being $(x_i, y_i, z_i) \in \mathbb{R}^3$ in world space, $\mathbf{c}_i \in \mathbb{R}^3$ denotes the colour and $\mathbf{o}_i \in [0, 1]$ the vertex opacity. A triangular face $\mathbf{F}_{m}$ is defined with three vertices $\mathbf{F}_m = \{\mathbf{v}_i, \mathbf{v}_j, \mathbf{v}_k\}$. 
The topology is defined by a set of faces $\mathcal{F} = \{\mathbf{f}_1, \dots, \mathbf{f}_M\}$, where each face $\mathbf{f}_m = (i_m, j_m, k_m)$ is an ordered triplet of vertex indices specifying the triangle. This separation of geometry and topology allows for dynamic re-indexing, enabling online Delaunay triangulation when a connected mesh is required. 

The colour contribution to a pixel inside a triangle's 2D projection is obtained by interpolating the vertex colours using barycentric coordinates: $ \mathbf{c}_{\mathbf{F}_m}(\lambda_i, \lambda_j, \lambda_k) = \lambda_i \mathbf{c}_i + \lambda_j \mathbf{c}_j + \lambda_k \mathbf{c}_k$
where $\lambda_i + \lambda_j + \lambda_k = 1$ and $\lambda_i, \lambda_j, \lambda_k \ge 0$. 
Note that this vertex-shared parameterisation differs from that of Triangle Splatting \cite{held2025triangle} where each face $\mathbf{F}_{m}$ has a single colour $\mathbf{c}_m$. The mesh parameterisation enables feature sharing across adjacent triangles, so that changes to connectivity do not also introduce discontinuities in the vertex attributes.
In a departure from Mesh Splatting, which sets the face opacity to the minimum of its vertices ($\mathbf{o}_{\mathbf{F}_m} = \min(\mathbf{o}_i, \mathbf{o}_j, \mathbf{o}_k)$), we define the face opacity as the average of its constituent vertex opacities: $\mathbf{o}_{\mathbf{F}_m} = \frac{1}{3}(\mathbf{o}_i + \mathbf{o}_j + \mathbf{o}_k)$. The minimum operator is discontinuous and routes gradients solely to the vertex with the lowest opacity. By using the average, we ensure that all three point opacities receive gradients and are jointly optimised, which improves the map quality in our online system.

\subsection{Differentiable Rasterisation}

To ensure the rasterisation process is differentiable, Held et al. \cite{held2025triangle} introduce a 2D signed distance field $\phi$ of the triangle in image space. 
For each triangle $\mathbf{F}_m = \{\mathbf{v}_i, \mathbf{v}_j, \mathbf{v}_k\}$, the 3D vertices are projected into image space via:

\begin{equation}
    \mathbf{v}_I = \pi(\mathbf{T}_{CW} \mathbf{v}_W) 
    ~,
    \label{eq:projection}
\end{equation}

\noindent where $\mathbf{v}_W \in \{\mathbf{v}_i, \mathbf{v}_j, \mathbf{v}_k\}$ denotes a vertex in world space (expressed in homogeneous coordinates for the transformation), $\mathbf{T}_{CW} \in \mathbf{SE}(3)$ is the camera pose, and $\pi$ is the projection operator. The influence of a pixel $\mathbf{p}$ is then defined as a window function $I(\mathbf{p})$ based on the signed distance field (SDF) $\phi$ of the projected 2D face \cite{held2025triangle}. The SDF is given by the maximum of the linear edge functions:

\begin{equation}
    \phi(\mathbf{p}) = \max_{i \in \{1,2,3\}} L_i(\mathbf{p}), \quad \text{where} \quad L_i(\mathbf{p}) = \mathbf{n}_i \cdot \mathbf{p} + d_i 
    ~,
    \label{eq:sdf}
\end{equation}

\noindent Here, $\mathbf{n}_i$ are the outward-facing unit normals of the 2D edges, and $d_i$ are their corresponding offsets, computed directly from the projected vertices $\mathbf{v}_I^{(i)}$.  We omit the details of their computation for simplicity.
The window function is then defined relative to the face's incentre $\mathbf{s} \in \mathbb{R}^2$, the point where the SDF is minimised:
\begin{equation}
    I(\mathbf{p}) = \text{ReLU} \left( \frac{\phi(\mathbf{p})}{\phi(\mathbf{s})} \right)^\sigma 
    ~,
    \label{eq:window}
\end{equation}

\noindent where $\sigma > 0$ is a parameter controlling the smoothness of the function. In Triangle Splatting the $\sigma$ parameter is learnable per-triangle, whereas in Mesh Splatting it is a global constant shared across all triangles. The formulation ensures that the influence $I(\mathbf{p})$ is exactly $1$ at the incentre of the face, decays to $0$ at the face boundaries, and remains $0$ everywhere outside the face.  After projection the colour for each pixel $\mathbf{p}$ is accumulated in the following manner: 
\begin{equation}
\label{eq:alpha_compositing}
    C(\mathbf{p}) = \sum_{n=1}^{N} \mathbf{c}_{\mathbf{F}_n} \mathbf{o}_{\mathbf{F}_n} I_n(\mathbf{p}) \left( \prod_{i=1}^{n-1} (1 - \mathbf{o}_{\mathbf{F}_i} I_i(\mathbf{p})) \right),
\end{equation}
where the faces are sorted in depth order.

\subsection{Analytic Camera Pose Jacobian}

Camera pose optimisation typically requires $\sim$80 iterations of gradient descent to converge. Computing gradients via automatic differentiation incurs substantial overhead, so we instead derive analytical pose Jacobians and evaluate them directly in the backward pass of the CUDA kernel, which significantly reduces tracking runtime. We represent pose updates as elements of $\mathfrak{se}(3)$ \cite{sola2018micro} and derive the Jacobian of the rendering loss with respect to the pose. Applying the chain rule through projection and the world-to-camera transformation we get:

\begin{equation}
\label{eq:loss_pose_chain}
\frac{\partial \mathcal{L}}{\partial \mathbf{T}_{CW}}
=
\frac{\partial \mathcal{L}}{\partial \mathbf{v}_I}
\frac{\partial \mathbf{v}_I}{\partial \mathbf{v}_C}
\frac{\mathcal{D} \mathbf{v}_C}{\mathcal{D} \mathbf{T}_{CW}} ,
\end{equation}

\noindent  where $\mathbf{v}_C$ denotes a vertex in camera space.
Taking partial derivatives on the manifold and using the same notation as \cite{sola2018micro} where $\tau \in \mathfrak{se}(3) $ we define:

\begin{equation}
    \frac{\mathcal{D}f(\boldsymbol{T})}{\mathcal{D}\boldsymbol{T}} \triangleq \lim_{\tau \to 0} \frac{\mathrm{Log}(f(\mathrm{Exp}(\tau) \circ \boldsymbol{T}) \circ f(\boldsymbol{T})^{-1})}{\tau},
\end{equation}

\noindent  And thus derive the following Jacobian of vertices with respect to the pose:

\begin{equation}
    \label{eq:dvdtau}
\frac{\mathcal{D}\mathbf{v}_C}{\mathcal{D} \mathbf{T}_{CW}} = \begin{bmatrix} \mathbf{I}_3 & \hspace{0.5mm}, \hspace{2mm} -[\mathbf{v}_C]_\times \end{bmatrix} ,
\end{equation}

\noindent  where $\mathbf{v}_\times$ denotes the skew symmetric matrix of the 3D vector.

\subsection{Tracking and Mapping}

We follow a sequential single-processed tracking and mapping framework where the camera pose is optimised to convergence after which the mapping module performs joint optimisation of camera poses and geometry. 

\subsubsection{Tracking}

The tracking frontend estimates the camera pose of the latest frame with respect to the map. This is achieved by minimising a joint objective comprising both photometric and geometric error. 
Incorporating an optimised Structural Similarity (SSIM) loss for more stable training \cite{taming3dgs}, we define $E_{pho}$ as a weighted combination of the per-pixel $\ell_1$ norm and the D-SSIM term between the image rendered from the triangles, $I(\mathcal{V}, \mathbf{T}_{CW})$, and the observed reference image $\bar{I}$:

\begin{equation}
    \label{eq:tracking_photometric_loss}
    E_{pho} = (1 - \lambda) \| I(\mathcal{V}, \mathbf{T}_{CW}) - \bar{I} \|_1 + \lambda \mathcal{L}_{\text{D-SSIM}}(I(\mathcal{V}, \mathbf{T}_{CW}), \bar{I}) ~.
\end{equation}
We also add a supervised depth tracking term. The depth loss $E_{dep}$ penalises the $\ell_1$ difference between the rendered depth map $D(\mathcal{V}, \mathbf{T}_{CW})$ and the observed depth $\bar{D}$:
\begin{equation}
\label{eq:tracking_depth_loss}
    E_{dep} = \| D(\mathcal{V}, \mathbf{T}_{CW}) - \bar{D} \|_1
~.
\end{equation}
The final tracking loss $E_{track}$ is formulated as the weighted sum of these terms, which is minimised to optimise the camera pose $\mathbf{T}_{CW} \in \mathbf{SE}(3)$:
\begin{equation}
    \label{eq:tracking_loss_total}
    E_{track} = E_{pho} + \lambda_{dep} E_{dep}
~.
\end{equation}

\subsubsection{Keyframing}

Keyframing is based on co-visibility checks, similar to MonoGS \cite{matsuki2024gaussian}. We keep track of triangles that were rendered from any keyframe during mapping, and a new keyframe is added based on the intersection-over-union of triangles with respect to the previous keyframe.  We maintain a keyframe buffer which also stores a count of triangles that were visible when that keyframe was rendered. Once a keyframe has been registered, we initiate the mapping phase.

\subsubsection{Mapping}

During mapping, we re-render from a selection of previous keyframes within the keyframe buffer and optimise both the keyframe poses and the map. We select 4 frames based on maximum covisibility, 2 random frames, and the most recently added keyframe. These are then replayed such that each keyframe is rendered for 30 iterations. The mapping loss $E_{map}$ is formulated to jointly optimise for photometric reconstruction and geometry.

\subsubsection{Photometric Supervision} The photometric error $E_{pho}$ penalises the difference between the rendered image $I(\mathcal{V}, \mathbf{T}_{CW})$ and the reference image $\bar{I}$. We use the same $E_{pho}$ from the earlier tracking phase.

\subsubsection{Geometric Supervision}

\begin{figure}[h]
    \centering
    \includegraphics[width=\linewidth]{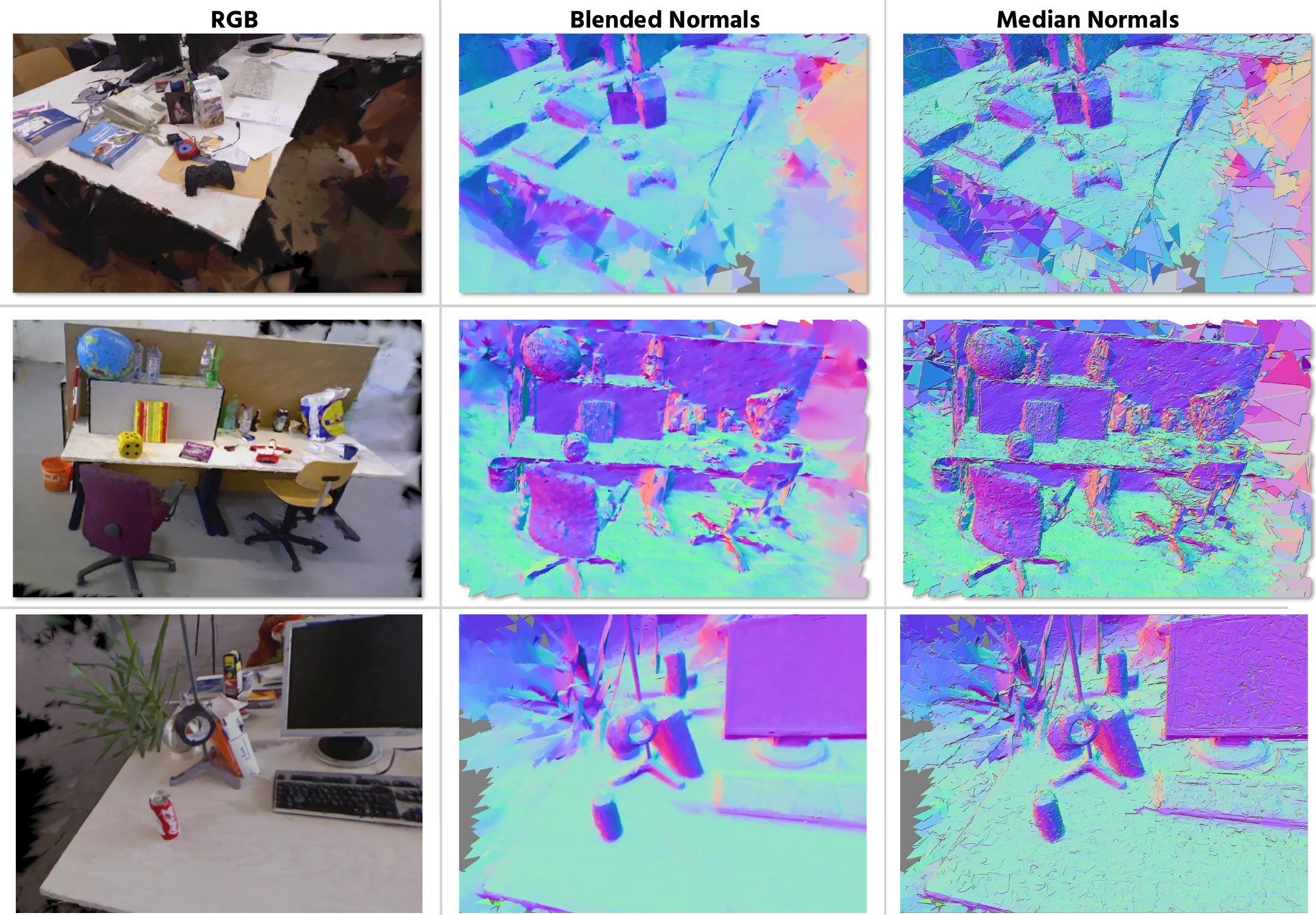}
    \caption{\textbf{Qualitative results of our method on the TUM RGB-D dataset}. \textbf{(left)}: RGB rendering of the reconstructed map. \textbf{(middle)}: Average normal direction of all triangles intersecting each camera ray. \textbf{(right)}: Normal direction of the triangle with highest opacity along each ray. Overall, our method produces photorealistic results and accurate surface reconstructions.}
    \label{fig:tum}
\end{figure}

Upon keyframing, we add new triangles to the scene by back-projecting RGB-D measurements and initialising an equilateral triangle around each point. A triangle is spawned with its points on a sphere of radius $r_i$, determined by the distance to its nearest neighbour, ensuring triangles are sized appropriately to the local point cloud density.

Following 4DTAM \cite{matsuki20254dtam}, we compute surface normal estimates directly from sensor depth measurements to initialise the triangle orientation. We compute these normal maps by taking finite differences of points in the back-projected depth. The sensor normals $\bar{\mathbf{N}}$ are thus formulated:
\begin{equation}
    \bar{\mathbf{N}} = \frac{\nabla_x \mathbf{p}_d \times \nabla_y \mathbf{p}_d}{\| \nabla_x \mathbf{p}_d \times \nabla_y \mathbf{p}_d \|}
    ~,
\end{equation}
where $\mathbf{p}_d$ denotes the back-projected depth points. These normal maps are stored within the keyframes and used as direct supervision during mapping. The normal regularisation loss $E_{norm}$ aligns the rendered geometry with the sensor data by penalising the cosine distance between the rendered normal map $N(\mathcal{V}, \mathbf{T}_{CW})$ returned by the rasteriser, and the pre-computed sensor normal map $\bar{\mathbf{N}}$ :

\begin{equation}
    \label{eq:normal_loss}
    E_{norm} = \sum_{\mathbf{p} \in \mathcal{P}} \left( 1 - N(\mathcal{V}, \mathbf{T}_{CW})_{\mathbf{p}}^T \bar{\mathbf{N}}_{\mathbf{p}} \right)
    ~,
\end{equation}

\noindent  where $\mathcal{P}$ denotes the set of all valid pixels. Additionally, we apply a depth loss $E_{dep}$ that penalises the $\ell_1$ difference between the rendered depth map $D(\mathcal{V}, \mathbf{T}_{CW})$ and the reference depth $\bar{D}$, further encouraging geometrically accurate reconstructions:
\begin{equation}
    \label{eq:depth_loss}
    E_{dep} = \| D(\mathcal{V}, \mathbf{T}_{CW}) - \bar{D} \|_1
~.
\end{equation}

\noindent We also introduce a regularisation term $E_{equi}$ to prevent the formation of degenerate, elongated triangles. For each face $\mathbf{f}_m \in \mathcal{F}$, we calculate the cosine of its three internal angles $\theta_{m,1}, \theta_{m,2}, \theta_{m,3}$ via the dot product of its normalised edge vectors. The loss penalises the difference between these cosines and $\cos(60^\circ) = 0.5$:

\begin{equation}
\label{eq:equilateral_loss}
    E_{equi} = \frac{1}{|\mathcal{F}|} \sum_{\mathbf{f}_m \in \mathcal{F}} \frac{1}{3} \sum_{n=1}^{3} \left( \cos \theta_{m,n} - 0.5 \right)^2
~.
\end{equation}

\vspace{2mm}

\noindent The total mapping loss is the weighted sum of these visual, geometric, and regularisation objectives:

\begin{equation}
     \label{eq:mapping_loss}
     E_{map} = E_{pho} + \lambda_{dep} E_{dep} + \lambda_{norm} E_{norm} + \lambda_{equi} E_{equi}
     ~.
\end{equation}

\subsubsection{Pruning} 

At every keyframe, unstable or redundant triangles are pruned from the map. This step is important to ensure the Delaunay triangulation does not contain geometrically inconsistent triangles, or those with low opacity in the blended render. A triangle is marked for removal if its mean vertex opacity falls below a threshold $\epsilon_{o}$, or if its 
projected image-space area exceeds a maximum size $\epsilon_{a}$, preventing 
large, under-resolved triangles from persisting in the map.

\begin{figure}[ht!]
    \centering
    \includegraphics[width=\linewidth]{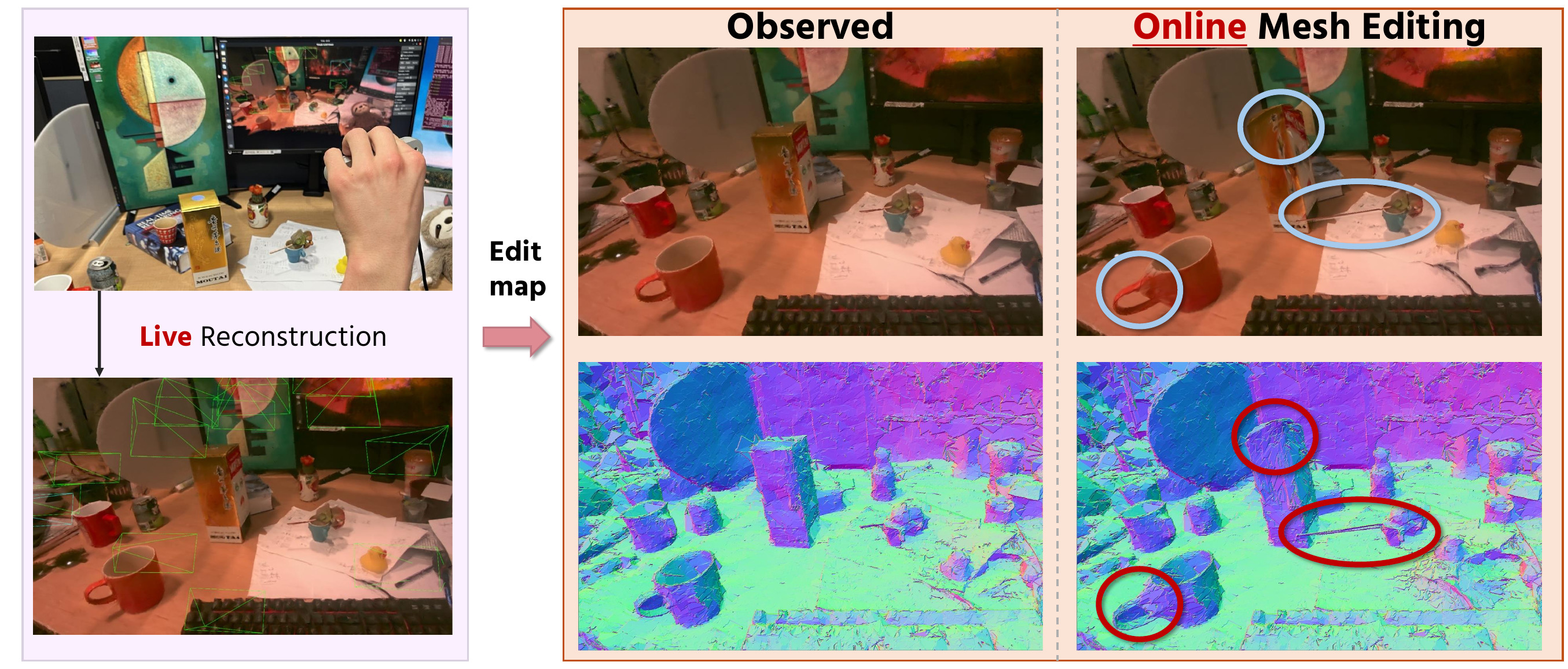}
    \caption{\textbf{Online mesh editing}. The triangle mesh representation supports online editing after Delaunay triangulation. Edited regions are circled in both RGB and surface normal renderings. Notably, our method handles thin structures such as the chameleon's tongue and the mug's handle \textbf{(circled)}. It also maintains realistic appearance which differs from methods which use TSDF Fusion.}
    \label{fig:edit}
\end{figure}

\subsubsection{Densification}

Primitives are added also during the mapping phase using the refinement strategy proposed by Fang et al. \cite{fang2024mini} where triangles whose projected pixel coverage exceeds a blur threshold 
$\theta_{blur} \cdot H \cdot W$ are selected as split candidates, to reduce blurry regions. Only 
triangles that are visible in the current replay window are eligible for densification.

We perform triangle splitting using the strategy proposed in \cite{Loop1987SmoothSS} where each selected primitive is subdivided into four smaller ones by 
introducing midpoint vertices along each edge. The positions, colours, and opacities of 
the three midpoint vertices are initialised by linear interpolation between 
their respective parent vertices. The original triangles are then removed and 
replaced by their four children. While Mesh Splatting \cite{Held2026MeshSplatting} proposed sharing vertices between triangles, even during the triangle soup stage of training, we found that this constrains points during optimisation and leads to poor convergence.

\section{Experiments}

\subsection{Implementation Details}

We run all experiments on a desktop PC with a single NVIDIA GeForce RTX 4090 GPU and AMD Ryzen 9 9950X 16-Core 4.3 GHz CPU. We report all results with a single processed implementation which performs mapping and tracking sequentially. The rasterisation pipeline is written in CUDA/C++ while the main SLAM loop was developed in PyTorch.

\begin{table}[h]
\centering
\caption{ATE (cm) RGB-D camera tracking results on TUM.}
\label{tab:tracking_results}
\setlength{\tabcolsep}{5pt} 
\resizebox{\textwidth}{!}{%
\begin{tabular}{lllcccc}
\toprule
\textbf{Input} & \textbf{Loop Closure} & \textbf{Method} & \textbf{fr1/desk} & \textbf{fr2/xyz} & \textbf{fr3/office} & \textbf{Avg.} \\ \midrule
\multirow{13}{*}{RGB-D} & \multirow{9}{*}{w/o} & iMAP & 4.90 & 2.00 & 5.80 & 4.23 \\
 &  & NICE-SLAM & 4.26 & 6.19 & 3.87 & 4.77 \\
 &  & DI-Fusion & 4.40 & 2.00 & 5.80 & 4.07 \\
 &  & Vox-Fusion & 3.52 & 1.49 & 26.01 & 10.34 \\
 &  & Co-SLAM & 2.40 & 1.70 & 2.40 & 2.17 \\
 &  & Point-SLAM & 4.34 & 1.31 & 3.48 & 3.04 \\
 &  & MonoGS & \textbf{1.50} & 1.44 & \textbf{1.49} & \textbf{1.47} \\
&  & MonoGS-2D & \underline{1.58} & \underline{1.20} & \underline{1.83} & \underline{1.54} \\
 &  & \textbf{Ours} & 1.77 & \textbf{1.12} & \underline{1.83} & 1.57 \\ \cmidrule{2-7} 
 & \multirow{3}{*}{w/} & BAD-SLAM & 1.70 & 1.10 & 1.70 & 1.50 \\
 &  & Kintinous & 3.70 & 2.90 & 3.00 & 3.20 \\
 &  & ORB-SLAM2 & \textbf{1.60} & \textbf{0.40} & \textbf{1.00} & \textbf{1.00} \\ \bottomrule
\end{tabular}%
}
\end{table}

\subsection{Datasets \& Baseline}

We benchmark our method using both TUM and Replica datasets following \cite{sucar2021imap, matsuki2024gaussian}.
We compare with similar map-centric RGB-D SLAM methods with explicit and implicit representations. For depth and Chamfer distance we compare with Gaussian Splatting SLAM (MonoGS) and our re-implementation of MonoGS using 2D Gaussian Splatting \cite{huang20242d}, which we label MonoGS-2D*. We also compare with the results presented in 4DTAM \cite{matsuki20254dtam}. To extract a mesh with TSDF Fusion we follow the same procedure proposed in \cite{huang20242d}. We also compare to a `pruned' version of our Delaunay mesh, where we remove triangles that are occluded in all training views.  
\subsection{Quantitative Results}

We report Root Mean Squared Error (RMSE) for the Absolute Tracking Error (ATE) of keyframes in   \cref{tab:tracking_results} and  \cref{tab:replica_tracking}.  L1 depth is shown in \cref{tab:depth_l1_replica} and Chamfer Distance in \cref{tab:replica_eval_chamf_time} on the Replica dataset. In \cref{tab:replica_eval_chamf_time} we also provide the mesh generation times of the different methods. We achieve comparable ATE in both datasets and outperform the baselines on geometry.

While MonoGS-2D* achieves the most accurate pose estimation and mean L1 depth error, this does not result in more accurate geometry from a 3D perspective.  The mean depth error reveals most frames with low error, but inaccurate depth maps exist in areas that are not viewed often.  While this is not enough to skew the mean depth metrics, after performing TSDF fusion, the Chamfer distance degrades in these areas.  Measuring the Chamfer distance also more directly measures the accuracy and utility of dense geometry.  For many tasks, achieving a consistent and accurate 3D reconstruction is of greater importance than depth estimation, so we highlight the ability of our method to achieve the best upper bound using TSDF reconstruction, while also providing a more flexible and photorealistic reconstruction using Delaunay triangulation. In this way, our method is particularly well-suited for on-the-fly reconstruction where a mesh is required at any given moment, as evidenced by the live demonstrations in the supplementary video and illustrated in \cref{fig:edit}.

\begin{table}[t]
\centering
\caption{ATE (cm) RGB-D camera tracking results on Replica.}
\label{tab:replica_tracking}
\setlength{\tabcolsep}{6pt} 
\resizebox{\textwidth}{!}{%
\begin{tabular}{lccccccccc}
\toprule
\textbf{Method} & \textbf{r0} & \textbf{r1} & \textbf{r2} & \textbf{o0} & \textbf{o1} & \textbf{o2} & \textbf{o3} & \textbf{o4} & \textbf{Avg.} \\ \midrule
iMAP \cite{sucar2021imap}      & 3.12 & 2.54 & 2.31 & 1.69 & 1.03 & 3.99 & 4.05 & 1.93 & 2.58 \\
NICE-SLAM \cite{zhu_niceslam_2022}   & 0.97 & 1.31 & 1.07 & 0.88 & 1.00 & 1.06 & 1.10 & 1.13 & 1.07 \\
Vox-Fusion \cite{yang2022vox}  & 1.37 & 4.70 & 1.47 & 8.48 & 2.04 & 2.58 & 1.11 & 2.94 & 3.09 \\
ESLAM \cite{johari-et-al-2023} & 0.71 & 0.70 & 0.52 & 0.57 & 0.55 & 0.58 & 0.72 & 0.63 & 0.63 \\
Point-SLAM \cite{sandstrom_point_2023} & 0.61 & 0.41 & 0.37 & 0.38 & 0.48 & 0.54 & 0.69 & 0.72 & 0.53 \\ \midrule
MonoGS \cite{matsuki2024gaussian}         & 0.44 & \underline{0.32} & 0.31 & 0.44 & 0.52 & 0.23 & 0.17 & 2.25 & 0.58 \\
MonoGS (sp)     & \textbf{0.33} & \textbf{0.22} & \underline{0.29} & 0.36 & \underline{0.19} & 0.25 & \textbf{0.12} & 0.81 & \underline{0.32} \\
MonoGS-2D   \cite{matsuki20254dtam}    & 0.42 & 0.43 & 0.35 & \underline{0.19} & \underline{0.19} & \underline{0.22} & 0.27 & 0.80 & 0.36 \\
MonoGS-2D*   & 0.55  & \underline{0.32} & 0.34  & \textbf{0.15}  & \textbf{0.10} & \textbf{0.09} & \underline{0.14} & \textbf{0.22} & \textbf{0.24} \\
\midrule
\textbf{Ours}   & \underline{0.36} & 0.41 & \textbf{0.25} & 0.32 & 0.30 & 0.45 & 0.73 & \underline{0.55} & 0.34 \\ \bottomrule
\end{tabular}%
}
\end{table}

\begin{figure}[h!]
    \centering
    \includegraphics[width=\linewidth]{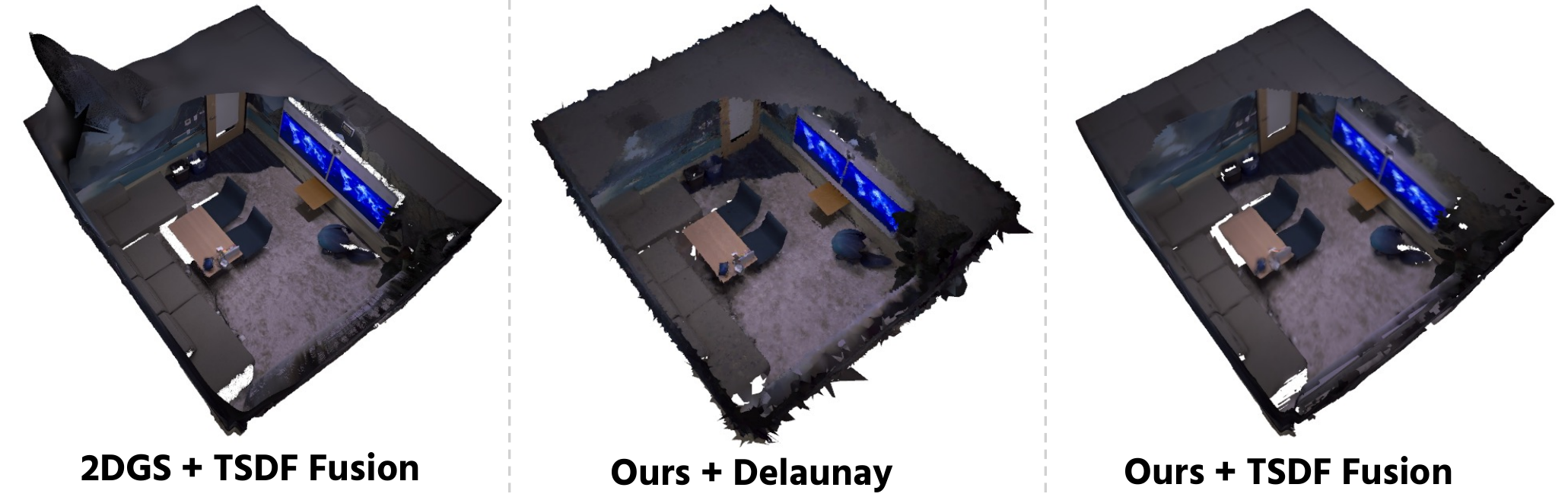}
    \caption{\textbf{Qualitative results on Replica}. Comparison of three methods ordered by ascending reconstruction accuracy (left to right). Our method with TSDF fusion achieves the highest quality; our method with Delaunay triangulation offers the best accuracy-speed trade-off while also enabling novel capabilities.}
    \label{fig:replica}
\end{figure}

\cref{fig:replica} shows a comparison between the meshes generated by MonoGS-2D* and those generated by our method using Delaunay and TSDF Fusion. While Delaunay meshing produces some irregular surfaces in unobserved regions, it more faithfully reconstructs the given input sequence as the surface is directly supervised during training. MonoGS-2D* notably produces artifacts in areas with few observations and depth outliers, which are not present in the TSDF meshes produced by our depth maps. This is reflected in \cref{tab:replica_eval_chamf_time} where our method achieves better Chamfer distance in all scenes for both Delaunay and TSDF Fusion of depth maps.

\begin{table}[t]
\centering
\caption{Depth L1 error (cm) on the Replica dataset $\downarrow$. }
\label{tab:depth_l1_replica}
\setlength{\tabcolsep}{6pt}
\resizebox{\textwidth}{!}{%
\begin{tabular}{lccccccccc}
\toprule
\textbf{Method} & \textbf{r0} & \textbf{r1} & \textbf{r2} & \textbf{o0} & \textbf{o1} & \textbf{o2} & \textbf{o3} & \textbf{o4} & \textbf{Avg.} \\ \midrule
MonoGS      & 3.00 & 3.47 & 4.66 & 3.10 & 6.08 & 6.15 & 4.77 & 4.94 & 4.52 \\
MonoGS-2D*   & \underline{1.44}  & \underline{0.78} & \textbf{0.80}  & \textbf{0.44}  & \underline{0.36} & \underline{0.59} & \textbf{0.87} & \textbf{0.68} & \underline{0.74} \\ \midrule
\textbf{Ours} & \textbf{0.77} & \textbf{0.70} & \underline{0.91}  & \underline{0.48} & \textbf{0.33} & \textbf{0.58}& \underline{0.95} & \underline{0.74} & \textbf{0.68} \\ \bottomrule
\end{tabular}%
}
\end{table}

\begin{table}[t]
\centering
\caption{Quantitative evaluation and computational trade-offs on the Replica dataset. We report both reconstruction quality (Chamfer Distance in cm $\downarrow$, evaluated with 1M samples) and mesh generation time (wall clock in seconds $\downarrow$). For TSDF extraction, we render depth maps from MonoGS (sp), MonoGS-2D*, and our representation before performing fusion. We compare this against running Restricted Delaunay triangulation on the triangle soup. TSDF Fusion on our depth maps achieves the highest geometric accuracy, however the pruned Delaunay triangulation offers highly competitive quality at less than half the computational time. Best results are \textbf{bolded} and second best are \underline{underlined} within their respective sections.}
\label{tab:replica_eval_chamf_time}
\setlength{\tabcolsep}{6pt}
\resizebox{\textwidth}{!}{%
\begin{tabular}{lccccccccc}
\toprule
\textbf{Method} & \textbf{r0} & \textbf{r1} & \textbf{r2} & \textbf{o0} & \textbf{o1} & \textbf{o2} & \textbf{o3} & \textbf{o4} & \textbf{Avg.} \\ \midrule
\multicolumn{10}{c}{\textit{Reconstruction Quality: Chamfer Distance (cm) $\downarrow$}} \\ \midrule
MonoGS + TSDF                   & 3.76 & 4.39 & 4.63 &  2.93 & 4.78 &  4.18  & 4.36 & 3.26 & 4.03 \\
MonoGS-2D* + TSDF               & 1.93 & 1.16 & 1.54 & 1.56 & 0.70 & 1.49 & 1.37 & 1.15 & 1.36 \\ 
\hline 
\textbf{Ours + TSDF}            & \textbf{1.00} & \textbf{0.77} & \textbf{1.01} & \textbf{0.66} & \textbf{0.56} & \textbf{1.12} & \textbf{1.69} & \textbf{0.78} & \textbf{0.95} \\ 
\textbf{Ours + Delaunay}        & 1.99 & 1.35 & 1.41 & 1.23 & 1.18 & 1.55 & 2.38 & 1.28 & 1.55 \\ 
\textbf{Ours + Delaunay (pruned)}& \underline{1.16} & \underline{1.01} & \underline{1.09} & \underline{0.77} & \underline{0.69} & \underline{1.37} & \underline{2.00} & \underline{1.02} & \underline{1.14} \\ \midrule
\multicolumn{10}{c}{\textit{Mesh Generation Time (seconds) $\downarrow$}} \\ \midrule
\textbf{Ours + TSDF}            & 47.40 & 25.10 & 32.20 & 25.60 & 13.50 & 33.90 & 50.20 & 39.60 & 33.44 \\ 
\textbf{Ours + Delaunay}        & \textbf{9.04} & \textbf{11.63} & \textbf{13.19} & \textbf{9.19} & \textbf{8.37} & \textbf{12.53} & \textbf{10.36} & \textbf{15.10} & \textbf{11.18} \\ 
\textbf{Ours + Delaunay (pruned)}& \underline{13.05} & \underline{15.92} & \underline{18.33} & \underline{13.34} & \underline{11.78} & \underline{17.80} & \underline{14.99} & \underline{20.10} & \underline{15.66} \\ \bottomrule
\end{tabular}%
}
\end{table}

\cref{tab:replica_eval_chamf_time} also highlights a trade-off between geometric accuracy and computational efficiency. Whilst TSDF fusion on our depth maps yields the lowest average Chamfer Distance (0.95 cm), it requires an average of 33.44 seconds to process a scene. In contrast, pruned Delaunay triangulation provides an effective middle ground. It achieves an average Chamfer Distance of 1.14 cm, remaining competitive with the TSDF upper bound and strictly outperforms MonoGS-2D* (1.36 cm). The mesh generation time of the pruned Delaunay method is less than half that of  TSDF fusion (15.66 seconds versus 33.44 seconds). The unpruned Delaunay variant is faster (11.18 seconds) at the cost of slightly increased geometric error (1.55cm), due to the retention of extraneous faces in unobserved or poorly constrained regions. Ultimately, this demonstrates the flexibility of our representation: it provides the capacity for reconstruction via TSDF, and simultaneously enables on-the-fly mesh generation via Delaunay triangulation without sacrificing significant geometric accuracy. We provide further analysis of these extraction methods, including performance across varying TSDF voxel resolutions, in the supplementary material.

\subsection{Runtime and Scaling Analysis}

\begin{table}[h]
      \centering
      \setlength{\tabcolsep}{10pt}
      \vspace{5mm}
      \caption{TUM and Replica latency and memory usage.}
      \label{tab:tum-results}
      \resizebox{0.9\columnwidth}{!}{%
      \begin{tabular}{lcccc}
      \toprule
       & \textbf{fr1/desk} & \textbf{fr2/xyz} & \textbf{fr3/office} & \textbf{office1} \\
      \midrule
      FPS $\uparrow$                & 0.82  & 2.33  & 1.29  & 0.55   \\
      Latency (ms) $\downarrow$     & 1225  & 429   & 775   & 1809   \\
      Frontend (ms)                 & 584   & 352   & 494   & 892    \\
      Backend (ms)                  & 642   & 77    & 282   & 918    \\
      Triangles                     & 29.0k & 24.0k & 34.4k & 152.2k \\
      Model size (MB) $\downarrow$  & 5.0   & 4.4   & 7.6   & 16.4   \\
      Peak GPU usage (GB)           & 0.47  & 0.47  & 0.53  & 1.25   \\
      \bottomrule
      \end{tabular}%
      }
\end{table}

We provide a runtime breakdown and system scaling analysis with map size in \cref{tab:tum-results}. We report per-scene FPS, \emph{average} per-frame latency, number of triangles and peak GPU usage with all components enabled across the TUM sequences and one Replica scene. The system runs live and interactively as shown in our video demonstration. The analysis shows that latency is generally correlated with map size, however sequences which keyframe less often can run much faster (e.g. fr2/xyz) as they perform fewer mapping iterations. We measure the map size using the model checkpoints, which indicate a low memory footprint. 

\subsection{Qualitative Capabilities}

The primary advantage of differentiable triangles as a representation lies in its capacity to provide both realistic rendering and a connected mesh, ensuring its immediate applicability to downstream tasks. To highlight this, we demonstrate live reconstruction that directly facilitates interactive, mesh-based scene editing. Whilst \cref{fig:edit} illustrates just one example of this capability, our supplementary video provides a comprehensive demonstration of the system in motion. This combination of visual fidelity and explicitly connected geometry is highly valuable for augmented reality experiences and collaborative design. Crucially, our method can incrementally update the mesh topology during the mapping phase, enabling complex, connectivity-based operations, including structural deformations and physical collisions, without requiring a separate offline extraction step.

\section{Conclusion and Future Work}

We presented an RGB-D SLAM system using triangles as a unified representation for photorealistic rendering, camera localisation, and 3D mapping. Beyond accurate trajectory estimation and consistent geometry, our approach unlocks online mesh-based scene editing and potential for live interactive digital twins.

Several directions remain for future work. First, the topological quality of the reconstructed mesh could be improved by incorporating constraints that enforce manifoldness and prevent self-intersections during optimisation, producing meshes more suitable for applications such as rigging. Second, recent work on differentiable triangle soup optimisation \cite{tojo2026diffsoup} has demonstrated extreme map simplification while preserving visual fidelity; integrating such compression into our pipeline could substantially reduce the number of primitives without sacrificing reconstruction quality. 
To achieve hard real-time (30fps) we would add multi-processing, conjugate gradients \cite{pehlivan2026matrixfree} and CUDA optimisations \cite{hahlbohm2026fastergs}. We prioritise a unified representation for tracking, mapping and meshing, however sparse bundle adjustment would also accelerate camera pose estimation. 
Finally, extending the system to monocular RGB input by tuning or using a strong prior \cite{leroy2024grounding} would broaden its applicability to more challenging sequences and outdoor reconstruction. Ultimately, this work demonstrates differentiable triangles can be used as a powerful, versatile representation for real-time spatial computing.

\section*{Acknowledgements}

This research is funded by the EPSRC On-Sensor Computer Vision grant \\(EP/Y020499/1). We would like to thank Ignacio Alzugaray, Shinjeong Kim, Marwan Taher, Hidenobu Matsuki, Riku Murai and other members of the Robot Vision Group for insightful discussions.

\clearpage

\appendix

\setcounter{section}{0}
\setcounter{figure}{0}
\setcounter{table}{0}
\setcounter{equation}{0}
\renewcommand{\thesection}{\Alph{section}}
\renewcommand{\thefigure}{S\arabic{figure}}
\renewcommand{\thetable}{S\arabic{table}}
\renewcommand{\theequation}{S\arabic{equation}}

\bibliographystyle{splncs04}
\bibliography{main}

@String(PAMI  = {IEEE Trans. Pattern Anal. Mach. Intell.})

@String(CVPR  = {IEEE Conf. Comput. Vis. Pattern Recog.})

@String(ICCV  = {Int. Conf. Comput. Vis.})

@String(ECCV  = {Eur. Conf. Comput. Vis.})

@String(NeurIPS = {Adv. Neural Inform. Process. Syst.})

@String(ICRA  = {IEEE Int. Conf. Robot. Autom.})

@String(TOG   = {ACM Trans. Graph.})

@string{TRO = {{{IEEE} Transactions on Robotics ({T-RO})}} }

@string{ SIGGRAPH = {{Proceedings of {SIGGRAPH}}} }

@string{ SIGGRAPHASIA = {{Proceedings of {SIGGRAPH} Asia}} }

@string{RSS = {{Proceedings of Robotics: Science and Systems ({RSS})}} }

@string{ISMAR = {{Proceedings of the International Symposium on Mixed and Augmented Reality ({ISMAR})}} }

@String(TDV = {Int. Conf. 3D Vis.})

@String(IJRR = {Int. J. Robot. Res.})

@inproceedings{matsuki2024gaussian,
  title={{Gaussian Splatting {SLAM}}},
  author={Matsuki, Hidenobu and Murai, Riku and Kelly, Paul HJ and Davison, Andrew J},
  booktitle=CVPR,
  year={2024}
}

@inproceedings{held2025triangle,
  title={Triangle Splatting for Real-Time Radiance Field Rendering},
  author={Held, Jan and Vandeghen, Renaud and Deliege, Adrien and Hamdi, Abdullah and Rebain, Daniel and Giancola, Silvio and Cioppa, Anthony and Vedaldi, Andrea and Ghanem, Bernard and Tagliasacchi, Andrea and others},
  booktitle=TDV,
  year={2025}
}

@inproceedings{sucar2021imap,
  title={{iMAP}: Implicit Mapping and Positioning in Real-Time},
  author={Sucar, Edgar and Liu, Shikun and Ortiz, Joseph and Davison, Andrew J},
  booktitle=ICCV,
  year={2021}
}

@article{kerbl20233d,
    title={{3D} {Gaussian} Splatting for Real-Time Radiance Field Rendering},
  author={Kerbl, Bernhard and Kopanas, Georgios and Leimk{\"u}hler, Thomas and Drettakis, George},
  journal=TOG,
  year={2023},
}

@inproceedings{shen2021dmtet,
  title={{Deep Marching Tetrahedra}: a Hybrid Representation for High-Resolution {3D} Shape Synthesis},
  author={Tianchang Shen and Jun Gao and Kangxue Yin and Ming-Yu Liu and Sanja Fidler},
  year={2021},
  booktitle=NeurIPS
}

@article{guedon2025milo,
  author={Gu\'{e}don, Antoine and Gomez, Diego and Maruani, Nissim and Gong, Bingchen and Drettakis, George and Ovsjanikov, Maks},
  title={{MILo}: Mesh-In-the-Loop {Gaussian} Splatting for Detailed and Efficient Surface Reconstruction},
  year={2025},
  journal=TOG
}

@inproceedings{mildenhall2020nerf,
  title={{NeRF}: Representing Scenes as Neural Radiance Fields for View Synthesis},
  author={Mildenhall, Ben and Srinivasan, Pratul P and Tancik, Matthew and Barron, Jonathan T and Ramamoorthi, Ravi and Ng, Ren},
  booktitle=ECCV,
  year={2020},
}

@inproceedings{huang20242d,
  title={{2D} {Gaussian} Splatting for Geometrically Accurate Radiance Fields},
  author={Huang, Binbin and Yu, Zehao and Chen, Anpei and Geiger, Andreas and Gao, Shenghua},
  booktitle=SIGGRAPH,
  year={2024}
}

@article{yu2024gaussian,
  title={{Gaussian Opacity Fields}: Efficient Adaptive Surface Reconstruction in Unbounded Scenes},
  author={Yu, Zehao and Sattler, Torsten and Geiger, Andreas},
  journal=TOG,
  year={2024}
}

@inproceedings{barron2022mip,
  title={{Mip-NeRF 360}: Unbounded Anti-Aliased Neural Radiance Fields},
  author={Barron, Jonathan T and Mildenhall, Ben and Verbin, Dor and Srinivasan, Pratul P and Hedman, Peter},
  booktitle=CVPR,
  year={2022}
}

@inproceedings{govindarajan2025radiant,
  title={{Radiant Foam}: Real-Time Differentiable Ray Tracing},
  author={Govindarajan, Shrisudhan and Rebain, Daniel and Yi, Kwang Moo and Tagliasacchi, Andrea},
  booktitle=ICCV,
  year={2025}
}

@mastersthesis{Loop1987SmoothSS,
  title={Smooth Subdivision Surfaces Based on Triangles},
  author={Charles T. Loop},
  year={1987},
  school={University of Utah}
}

@inproceedings{fang2024mini,
  title={{Mini-Splatting}: Representing Scenes with a Constrained Number of {Gaussians}},
  author={Fang, Guangchi and Wang, Bing},
  booktitle=ECCV,
  year={2024},
}

@article{schops2019surfelmeshing,
  title={{SurfelMeshing}: Online Surfel-Based Mesh Reconstruction},
  author={Sch{\"o}ps, Thomas and Sattler, Torsten and Pollefeys, Marc},
  journal=PAMI,
  year={2019},
}

@inproceedings{kinectfusion,
  author={Newcombe, Richard A. and Izadi, Shahram and Hilliges, Otmar and Molyneaux, David and Kim, David and Davison, Andrew J. and Kohi, Pushmeet and Shotton, Jamie and Hodges, Steve and Fitzgibbon, Andrew},
  booktitle=ISMAR,
  title={{KinectFusion}: Real-Time Dense Surface Mapping and Tracking},
  year={2011},
}

@inproceedings{pointbasedfusionKeller,
  author={Keller, Maik and Lefloch, Damien and Lambers, Martin and Izadi, Shahram and Weyrich, Tim and Kolb, Andreas},
  title={Real-Time {3D} Reconstruction in Dynamic Scenes Using Point-Based Fusion},
  year={2013},
  booktitle=TDV,
}

@inproceedings{schops_badslam_2019,
  author={Sch{\"o}ps, Thomas and Sattler, Torsten and Pollefeys, Marc},
  booktitle=CVPR,
  title={{BAD SLAM}: Bundle Adjusted Direct {RGB-D} {SLAM}},
  year={2019},
}

@inproceedings{matsuki20254dtam,
  title={{4DTAM}: Non-rigid Tracking and Mapping via Dynamic Surface {Gaussians}},
  author={Matsuki, Hidenobu and Bae, Gwangbin and Davison, Andrew J},
  booktitle=CVPR,
  year={2025}
}

@inproceedings{mai2026radiance,
  title={Radiance Meshes for Volumetric Reconstruction},
  author={Mai, Alexander and Hedstrom, Trevor and Kopanas, George and Kontkanen, Janne and Kuester, Falko and Barron, Jonathan T.},
  booktitle=CVPR,
  year={2026}
}

@inproceedings{taming3dgs,
  author={Mallick, Saswat Subhajyoti and Goel, Rahul and Kerbl, Bernhard and Steinberger, Markus and Carrasco, Francisco Vicente and De La Torre, Fernando},
  title={Taming {3DGS}: High-Quality Radiance Fields with Limited Resources},
  year={2024},
  booktitle=SIGGRAPHASIA,
}

@article{mueller2022instant,
  author={Thomas M\"uller and Alex Evans and Christoph Schied and Alexander Keller},
  title={Instant Neural Graphics Primitives with a Multiresolution Hash Encoding},
  journal=TOG,
  year={2022},
}

@inproceedings{leroy2024grounding,
  title={Grounding image matching in {3D} with {MASt3R}},
  author={Leroy, Vincent and Cabon, Yohann and Revaud, J{\'e}r{\^o}me},
  booktitle=ECCV,
  year={2024},
}

@article{mur-artal_orb-slam2_2017,
  title={{ORB}-{SLAM2}: An Open-Source {SLAM} System for Monocular, Stereo, and {RGB}-{D} Cameras},
  journal=TRO,
  author={Mur-Artal, Ra\'{u}l and Tard\'{o}s, Juan D.},
  year={2017},
}

@inproceedings{curless_tsdf_1996,
  author={Curless, Brian and Levoy, Marc},
  title={A Volumetric Method for Building Complex Models from Range Images},
  year={1996},
  booktitle=SIGGRAPH,
}

@INPROCEEDINGS{whelan_kintinuous_2013,
  author={Whelan, Thomas and Johannsson, Hordur and Kaess, Michael and Leonard, John J. and McDonald, John},
  booktitle=ICRA,
  title={Robust Real-Time Visual Odometry for Dense {RGB-D} Mapping},
  year={2013},
}

@inproceedings{whelan_elasticfusion_2015,
  title={{ElasticFusion}: Dense {SLAM} Without a Pose Graph.},
  author={Whelan, Thomas and Leutenegger, Stefan and Salas-Moreno, Renato F and Glocker, Ben and Davison, Andrew J},
  booktitle=RSS,
  year={2015},
}

@article{niessner_hashing_2013,
  author={Nie\ss{}ner, Matthias and Zollh\"{o}fer, Michael and Izadi, Shahram and Stamminger, Marc},
  title={Real-time {3D} Reconstruction at Scale Using Voxel Hashing},
  year={2013},
  journal=TOG,
}

@inproceedings{klein_parallel_2007,
  author={Klein, Georg and Murray, David},
  booktitle=ISMAR,
  title={Parallel Tracking and Mapping for Small {AR} Workspaces},
  year={2007},
}

@inproceedings{sandstrom_point_2023,
  author={Sandstr{\"o}m, Erik and Li, Yue and Van Gool, Luc and R. Oswald, Martin},
  title={{Point-SLAM}: Dense Neural Point Cloud-based {SLAM}},
  booktitle=ICCV,
  year={2023}
}

@inproceedings{zhu_niceslam_2022,
  title={{NICE-SLAM}: Neural implicit scalable encoding for {SLAM}},
  author={Zhu, Zihan and Peng, Songyou and Larsson, Viktor and Xu, Weiwei and Bao, Hujun and Cui, Zhaopeng and Oswald, Martin R and Pollefeys, Marc},
  booktitle=CVPR,
  year={2022}
}

@inproceedings{keetha_splatam_2024,
  title={{SplaTAM}: Splat, Track \& Map {3D} {G}aussians for Dense {RGB-D SLAM}},
  author={Keetha, Nikhil and Karhade, Jay and Jatavallabhula, Krishna Murthy and Yang, Gengshan and Scherer, Sebastian and Ramanan, Deva and Luiten, Jonathon},
  booktitle=CVPR,
  year={2024}
}

@inproceedings{yan2024gs,
  title={{GS-SLAM}: Dense visual {SLAM} with {3D} {Gaussian} splatting},
  author={Yan, Chi and Qu, Delin and Xu, Dan and Zhao, Bin and Wang, Zhigang and Wang, Dong and Li, Xuelong},
  booktitle=CVPR,
  year={2024}
}

@inproceedings{yang2022vox,
  title={{Vox-Fusion}: Dense Tracking and Mapping with Voxel-based Neural Implicit Representation},
  author={Yang, Xingrui and Li, Hai and Zhai, Hongjia and Ming, Yuhang and Liu, Yuqian and Zhang, Guofeng},
  booktitle=ISMAR,
  year={2022},
}

@inproceedings{johari-et-al-2023,
  author={Johari, M. M. and Carta, C. and Fleuret, F.},
  title={{ESLAM}: Efficient Dense {SLAM} System Based on Hybrid Representation of Signed Distance Fields},
  booktitle=CVPR,
  year={2023},
}

@article{sola2018micro,
  title={A Micro Lie Theory for State Estimation in Robotics},
  author={Sola, Joan and Deray, Jeremie and Atchuthan, Dinesh},
  journal={arXiv preprint arXiv:1812.01537},
  year={2018}
}

@inproceedings{marchingcubes,
  author={Lorensen, William E. and Cline, Harvey E.},
  title={Marching Cubes: A High Resolution {3D} Surface Construction Algorithm},
  year={1987},
  booktitle=SIGGRAPH,
}

@inproceedings{tojo2026diffsoup,
  author={Tojo, Kenji and Bickel, Bernd and Umetani, Nobuyuki},
  title={{DiffSoup}: Direct Differentiable Rasterization of Triangle Soup for Extreme Radiance Field Simplification},
  booktitle=CVPR,
  year={2026}
}

@inproceedings{Held2026MeshSplatting,
  title={{MeshSplatting}: Differentiable Rendering with Opaque Meshes},
  author={Held, Jan and Son, Sanghyun and Vandeghen, Renaud and Rebain, Daniel and Gadelha, Matheus and Zhou, Yi and Cioppa, Anthony and Lin, Ming C. and Van Droogenbroeck, Marc and Tagliasacchi, Andrea},
  booktitle=CVPR,
  year={2026}
}

@inproceedings{pehlivan2026matrixfree,
  title={Matrix-free Second-order Optimization of {Gaussian} Splats with Residual Sampling},
  author={Pehlivan, Hamza and Boscolo Camiletto, Andrea and Foo, Lin Geng and Habermann, Marc and Theobalt, Christian},
  booktitle=TDV,
  year={2026},
}

@inproceedings{hahlbohm2026fastergs,
  title={Faster-{GS}: Analyzing and Improving {Gaussian} Splatting Optimization},
  author={Hahlbohm, Florian and Franke, Linus and Eisemann, Martin and Magnor, Marcus},
  booktitle=CVPR,
  year={2026}
}

@article{tao2025spires,
  title={The {Oxford} Spires Dataset: Benchmarking Large-Scale {LiDAR}-Visual Localisation, Reconstruction and Radiance Field Methods},
  author={Tao, Yifu and Mu{\~n}oz-Ba{\~n}{\'o}n, Miguel {\'A}ngel and Zhang, Lintong and Wang, Jiahao and Fu, Lanke Frank Tarimo and Fallon, Maurice},
  journal=IJRR,
  year={2025},
}

\clearpage

\section*{ \textbf{Supplementary Material}}

\section{Geometry Analysis}

\begin{figure}[h]
    \centering
    \begin{subfigure}{0.48\textwidth}
        \centering
        \includegraphics[width=\linewidth]{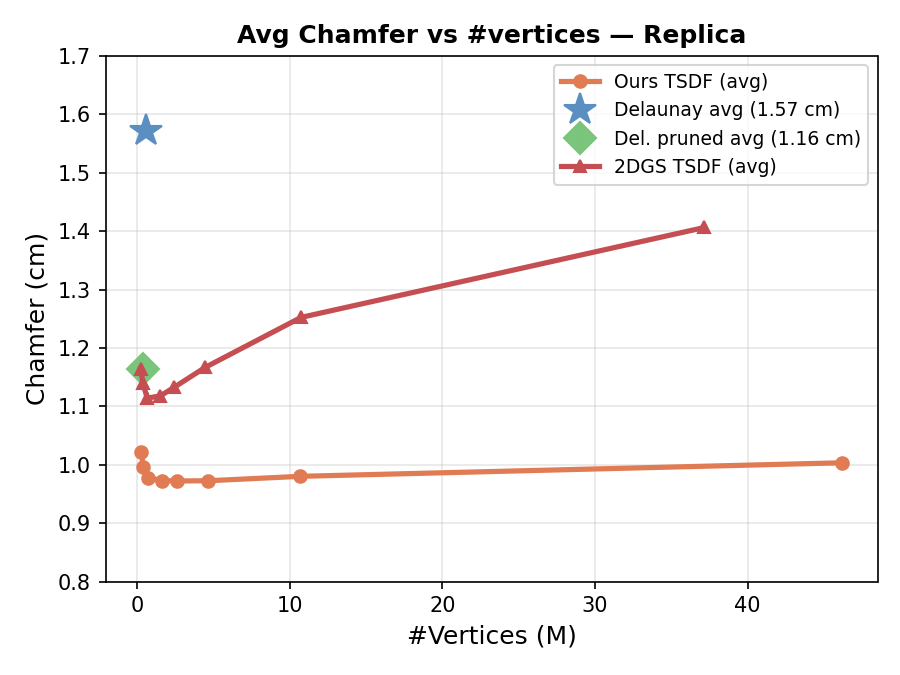}
        \caption{Chamfer vs. Number of Vertices}
    \end{subfigure}\hfill
    \begin{subfigure}{0.48\textwidth}
        \centering
        \includegraphics[width=\linewidth]{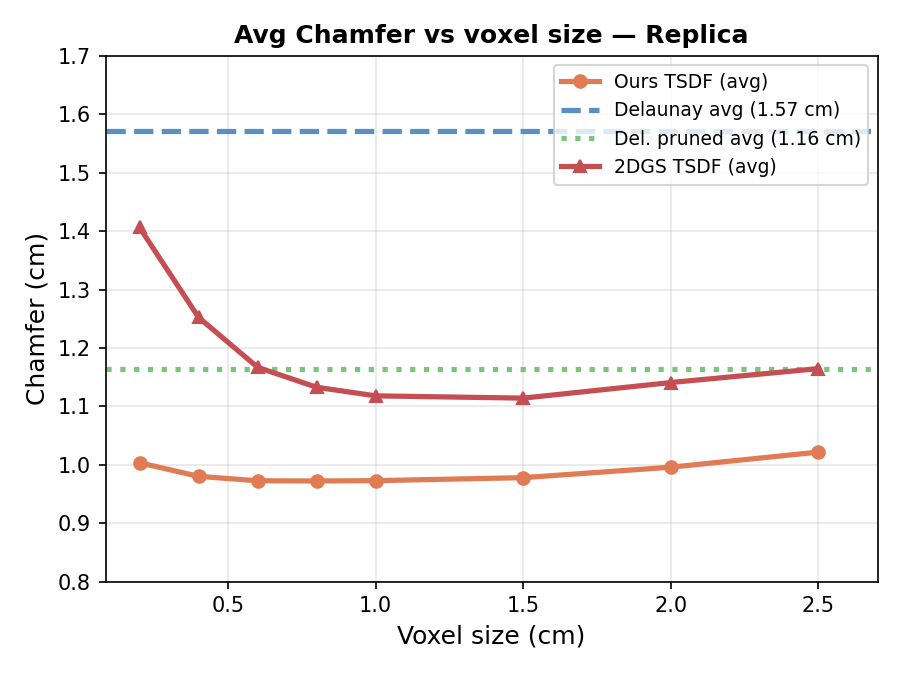}
        \caption{Chamfer vs. Voxel Size}
    \end{subfigure}
    
    \caption{Comparison of average Chamfer distances. Finer TSDF voxel sizes capture more noise, increasing the Chamfer error, whereas larger voxel sizes smooth the geometry at the cost of detail.}
    \label{fig:chamfer_comparisons}
\end{figure}

We provide further details on our geometry evaluation. The ground truth mesh is generated by running TSDF fusion with the ground truth poses and point maps provided by the Replica dataset using the exact extraction procedure proposed in \cite{huang20242d} (setting voxel size to $0.004$\,m and truncation threshold to $0.02$\,m). We then run TSDF fusion on the rendered depth maps derived from keyframes using the same configuration. The small voxel size means that the generation time is prohibitively slow for a real-time system. Furthermore, the fine granularity captures much of the noise present in the depth maps generated by MonoGS-2D*, which ultimately degrades the Chamfer distance, as illustrated in \cref{fig:chamfer_comparisons}. 

\begin{figure}[h]
    \centering
    \begin{subfigure}{0.323\textwidth}
        \centering
        \includegraphics[width=\linewidth]{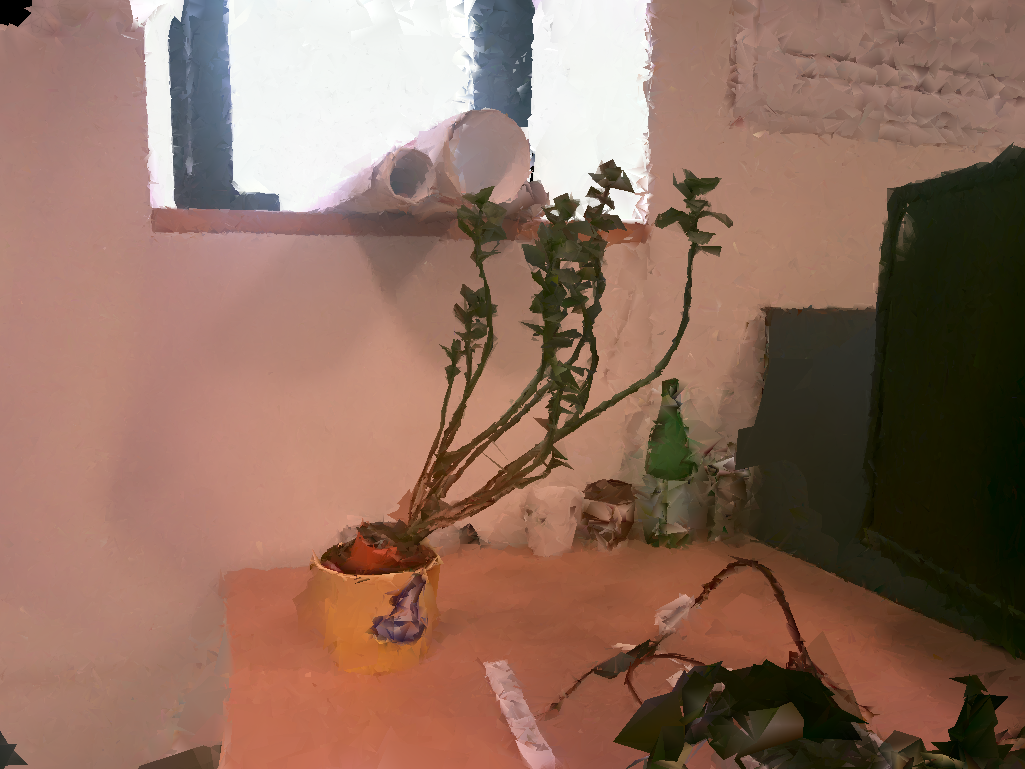}
        \caption{Delaunay (unpruned)}
        \label{fig:tree_tsdf1}
    \end{subfigure}%
    \hspace{0.005\textwidth}%
    \begin{subfigure}{0.323\textwidth}
        \centering
        \includegraphics[width=\linewidth]{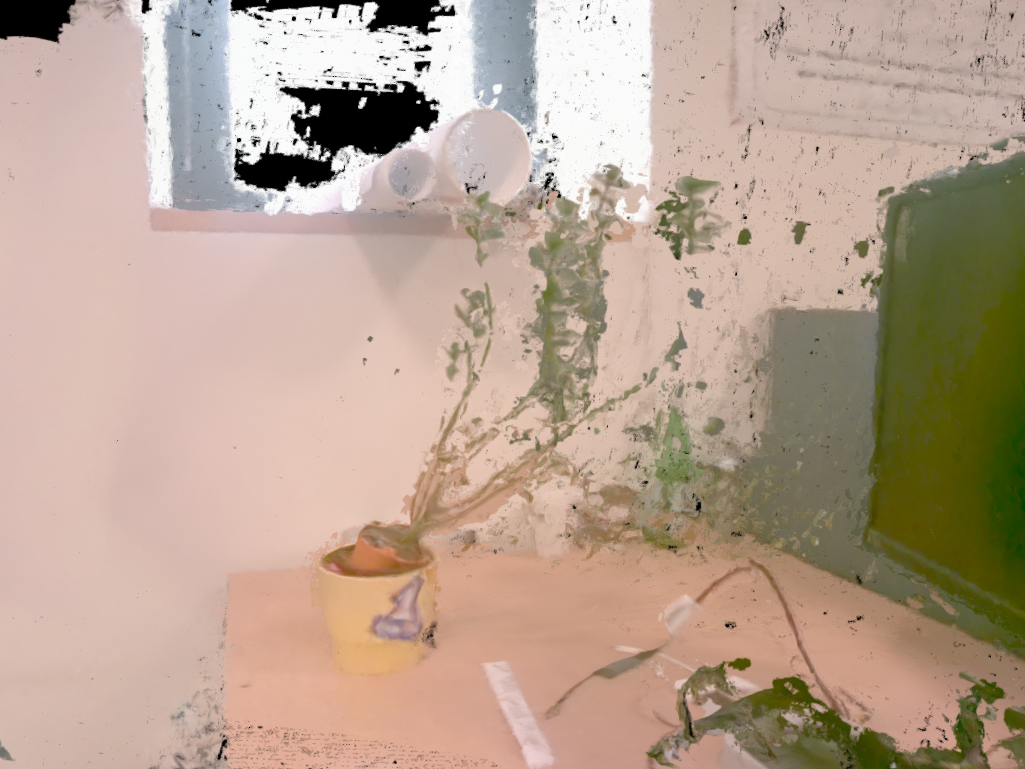}
        \caption{TSDF fusion (1mm voxel)}
        \label{fig:tree_tsdf1_o3d}
    \end{subfigure}%
    \hspace{0.005\textwidth}%
    \begin{subfigure}{0.323\textwidth}
        \centering
        \includegraphics[width=\linewidth]{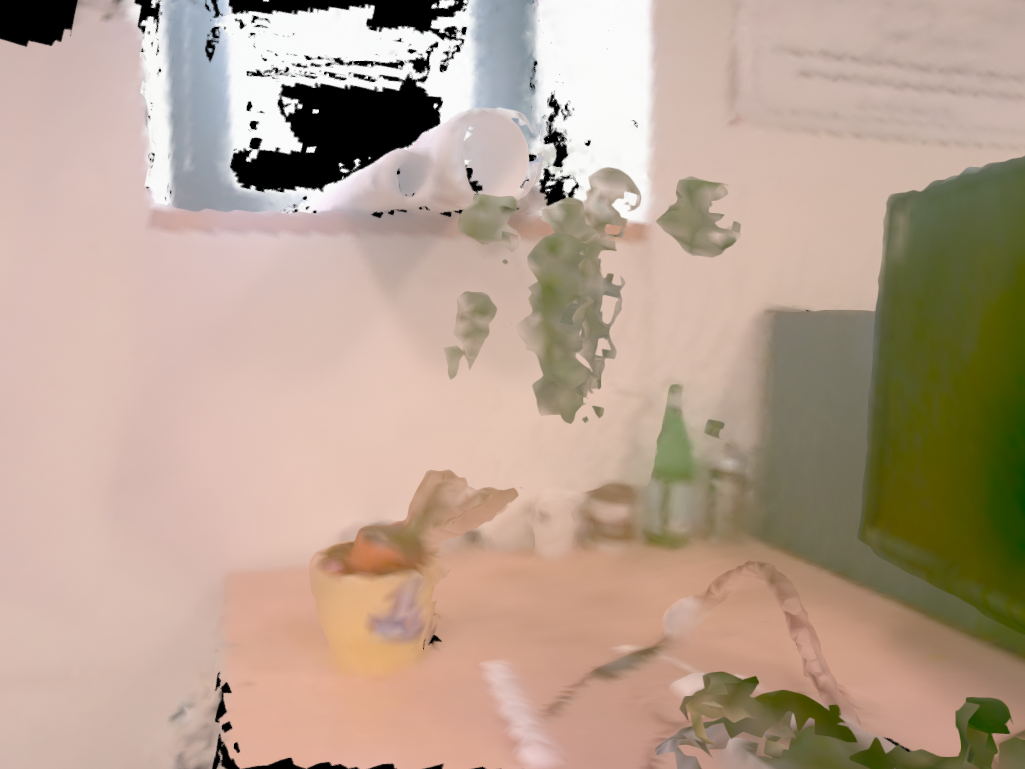}
        \caption{TSDF fusion (1cm voxel)}
        \label{fig:tree_tsdf2_1mm}
    \end{subfigure}
     \caption{Comparison of raw Delaunay mesh with TSDF fusion of different voxel sizes. At 1mm TSDF fusion has 23,098,956 verts, and 41,702,646 faces and at 1cm 356,037 verts, 495,685 faces. Note that much of the fine detail from the branches and leaves is lost. Delaunay has 156,318 verts and 594,658 faces and still retains these geometric details.}
    \label{fig:tree_comparisons}
    \vspace{-5mm}
\end{figure}

\begin{figure}[h]
    \centering
    \begin{subfigure}[b]{0.49\textwidth}
        \centering
        \includegraphics[width=\textwidth]{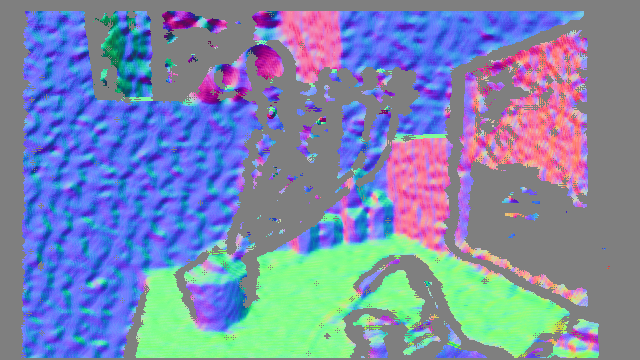}
        \caption{GT Normals}
        \label{fig:normals_gt}
    \end{subfigure}%
    \hspace{0.003\textwidth}%
    \begin{subfigure}[b]{0.49\textwidth}
        \centering
        \includegraphics[width=\textwidth]{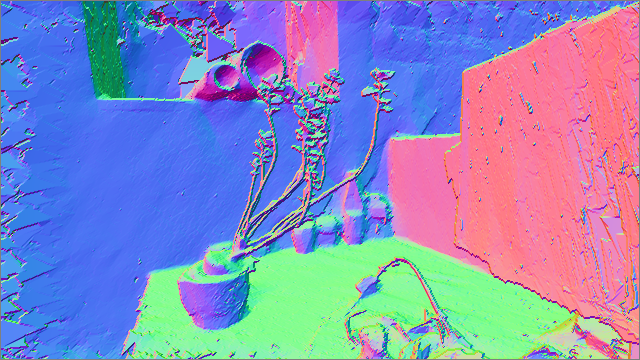}
        \caption{Our Reconstructed Normals}
        \label{fig:normals_ours}
    \end{subfigure}
    \caption{\textbf{Benefit of differentiable rendering.} Even with noisy/missing ground truth depth, photometric supervision successfully recovers precise geometric details such as the leaves and branches of the plant.}
    \label{fig:comparison_normals}
\end{figure}

Conversely, using a larger voxel size is much faster, smooths the geometry and removes outliers, yielding a better Chamfer score, but sacrifices fine mesh details, as shown in \cref{fig:tree_comparisons}. We benchmark this baseline against a raw Delaunay triangulation of the entire scene, as well as a pruned Delaunay mesh where triangles occluded across all keyframes are removed (approximating the outlier-removal effect of TSDF). We denote the latter as Delaunay (pruned).  We also demonstrate how using a combination of photometric supervision and depth yields far better geometry than depth alone.  \cref{fig:comparison_normals} shows the ground truth normal maps estimated from depth alongside our reconstruction.

We evaluate against TSDF fusion, however, this does not highlight a key contribution of our approach. TSDF fusion, like most prior reconstruction methods, generates a mesh that is separate from the representation used for tracking. In-the-loop applications such as online mesh deformation and physics simulation during tracking are therefore infeasible with these methods. Our system instead uses a single triangle representation for tracking, mapping, and meshing, unifying all of these downstream tasks with one representation. 

 Our system demonstrates robustness to real-world conditions (e.g. in TUM): incomplete depth, motion blur, and exposure changes. Low texture regions are challenging for all photometric SLAM systems, but depth supervision can increase robustness in these cases.
The plant deformation in our video demonstration shows the ability to recover high-fidelity connected geometry even with missing depth and partial occlusions around the branches.

\vspace{-5mm}

\vspace{-8mm}

\section{Timing Analysis}

\begin{figure}[ht]
    \centering
    \includegraphics[width=0.7\linewidth]{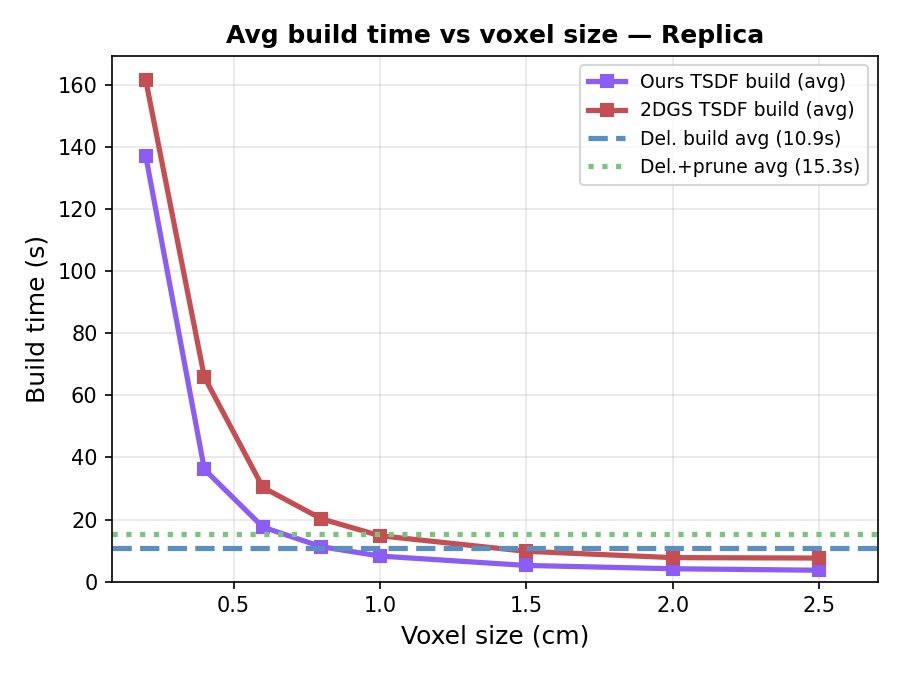}
    \caption{Average mesh generation time evaluated across different TSDF voxel sizes. Despite its theoretically higher asymptotic complexity, Delaunay triangulation is competitive even against coarse volumetric resolutions by exploiting scene sparsity.}
    \label{fig:buildtime}
\end{figure}

Larger voxel sizes significantly reduce mesh generation times for TSDF fusion, as shown in \cref{fig:buildtime}. However, even at these coarse resolutions, Delaunay triangulation remains competitive and fast enough to be usable for on-the-fly mesh generation. Despite having worse theoretical time complexity than volumetric TSDF fusion, using Delaunay triangulation exploits geometric sparsity and varying point densities across the scene. 

The cost of TSDF scales with total scene volume, while Delaunay with vertex density: visual quality is therefore significantly higher for differentiable rendering with Delaunay triangulation given similar vertex counts (see \cref{fig:tree_comparisons}), as triangles are adaptively allocated to detailed areas. 

We report the triangulation time for the whole scene, however in practice the ideal setup for on-the-fly Delaunay is to re-triangulate only the geometry within the current viewing frustum; localised updates are much faster and trivial in the vertex-parameterised setup, yet remain involved for a more global algorithm like TSDF fusion.

\section{Photometric Analysis}

\begin{figure}[ht]
    \centering
    \begin{subfigure}[b]{0.487\textwidth}
        \centering
        \includegraphics[width=\textwidth]{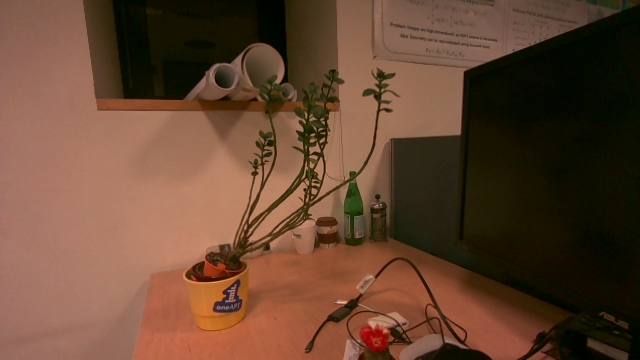}
        \caption{Ground Truth Reference}
        \label{fig:rgb_gt}
    \end{subfigure}%
    \hspace{0.003\textwidth}%
    \begin{subfigure}[b]{0.487\textwidth}
        \centering
        \includegraphics[width=\textwidth]{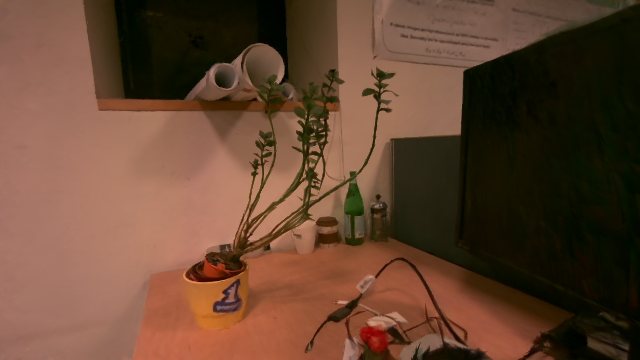}
        \caption{Our Reconstructed Output}
        \label{fig:rgb_ours}
    \end{subfigure}
    \caption{\textbf{Photorealistic Reconstruction.} Comparison showing that our method achieves high-fidelity view synthesis that closely matches the ground truth reference.}
    \label{fig:comparison_rgb}
\end{figure}

We also report PSNR/SSIM/LPIPS in \cref{tab:rendering_evaluation} on Replica. Photorealism is not the main focus of our method, however, it is comparable with other systems. Note that we do not do any optimisation of the triangle appearance at the end of the sequence. We calculate this in the same manner as \cite{matsuki2024gaussian}. We also show a side-by-side view in \cref{fig:comparison_rgb} to highlight the reconstruction quality of our system on real data.  Specularities can be modelled using higher order spherical harmonics (SH), however this increases parameter count significantly and is harder to train online. We opt for zeroth-order SH, as in comparable systems  (MonoGS).

\section{Outdoor scenes}

\begin{figure}[h]
    \centering
    \begin{subfigure}[b]{0.49\textwidth}
        \centering
        \includegraphics[width=\textwidth]{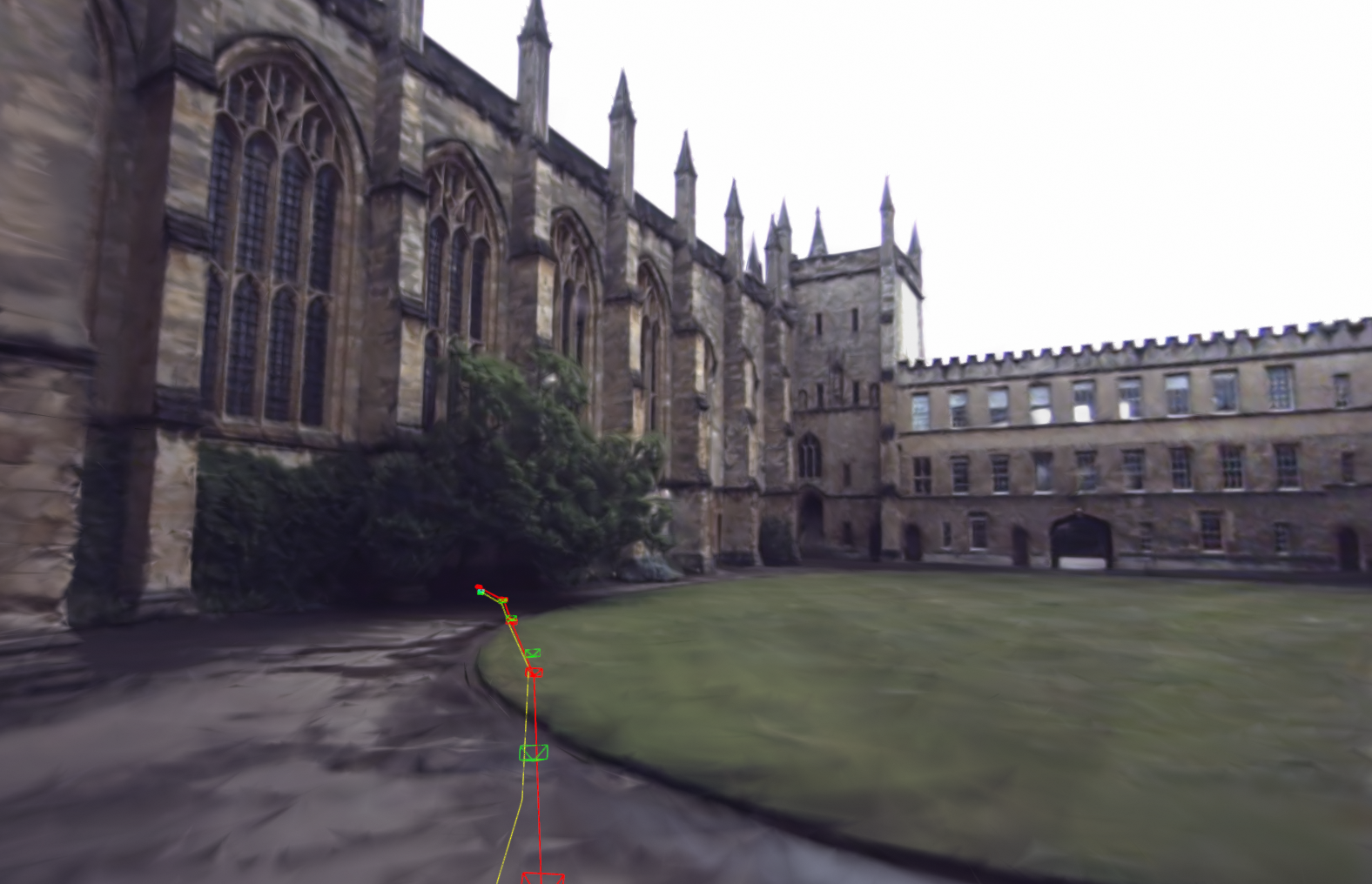}
        \caption{New College.}
        \label{fig:newcollege}
    \end{subfigure}%
    \hspace{0.003\textwidth}%
    \begin{subfigure}[b]{0.49\textwidth}
        \centering
        \includegraphics[width=\textwidth]{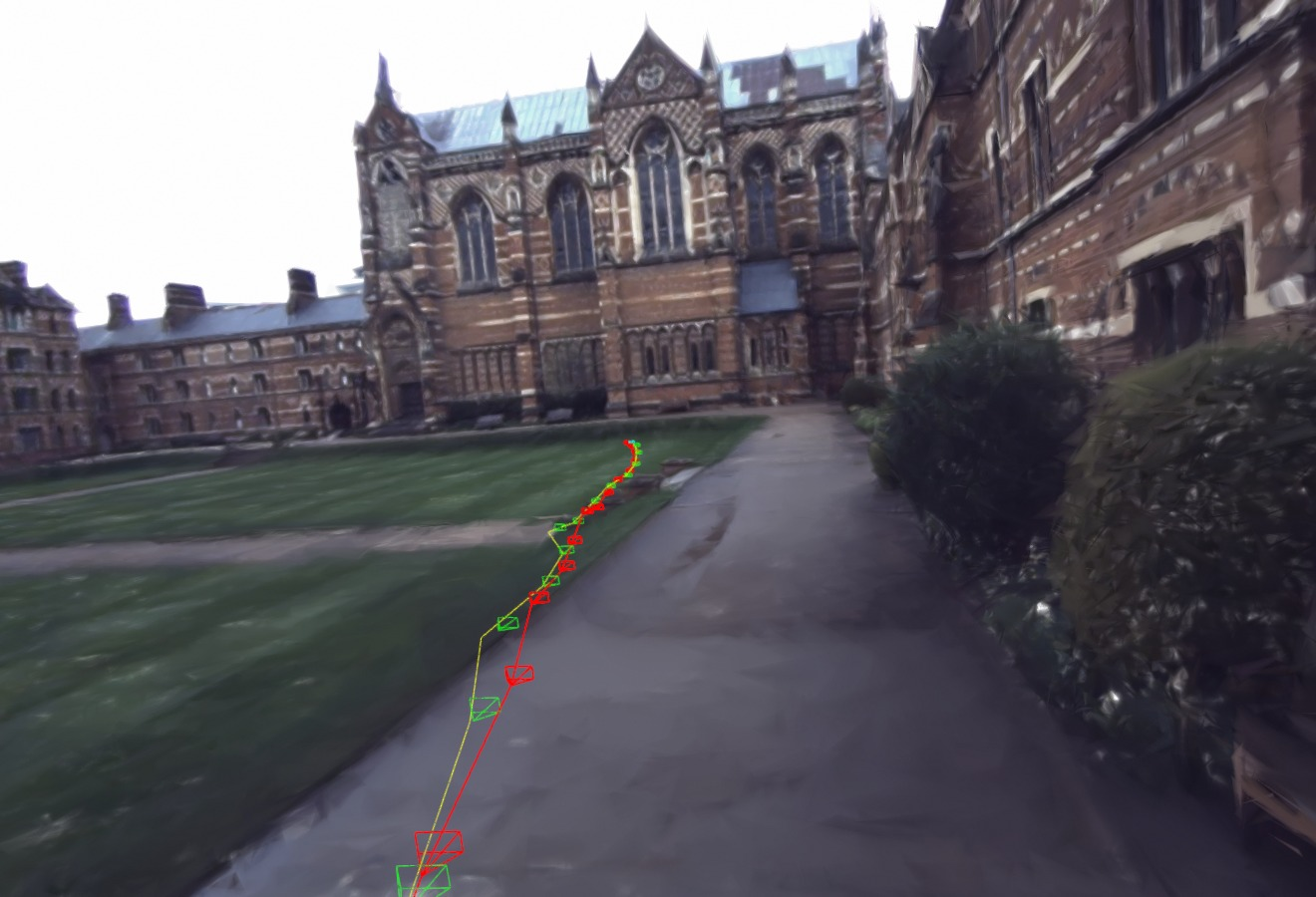}
        \caption{Keble College.}
        \label{fig:oxfordspires}
    \end{subfigure}
    \caption{Qualitative analysis on complex outdoor scenes, with monocular depth estimation.}
    \label{fig:outdoor}
\end{figure}

 We show our system generalises to monocular and large-scale outdoor settings in \cref{fig:outdoor}, on the Oxford Spires dataset \cite{tao2025spires}. We use MoGe-2 to estimate per-frame depth.
\cref{fig:outdoor} shows novel-view quality remains high, the trajectory follows the GT closely, and the vegetation is faithfully reconstructed (hedges, trees, lawn).  Loop-closure would be a valuable addition to correct for drift; however, the system  handles these large scale sequences without it. 

\begin{table}[h]
\centering
\caption{Quantitative rendering evaluation. We report PSNR, SSIM, and LPIPS.}
\label{tab:rendering_evaluation}
\setlength{\tabcolsep}{6pt}
\begin{tabular}{lccc}
\toprule
\textbf{Method} & \textbf{PSNR $\uparrow$} & \textbf{SSIM $\uparrow$} & \textbf{LPIPS $\downarrow$} \\ \midrule
NICE-SLAM [48]  & 24.42 & 0.809 & 0.233 \\
Vox-Fusion [45] & 24.41 & 0.801 & 0.236 \\
Point-SLAM [29] & \underline{35.17} & \textbf{0.975} & \underline{0.124} \\
MonoGS          & \textbf{38.94} & \underline{0.968} & \textbf{0.070} \\ \midrule
\textbf{Ours}   & 31.33 & 0.909 & 0.206 \\ \bottomrule
\end{tabular}
\end{table}

\vspace{-5mm}

\section{Ablation}

\cref{tab:ablation} presents an ablation study on the TUM fr1/desk sequence, where we independently remove each loss component from the mapping objective. Removing the SSIM loss degrades both tracking accuracy and rendering quality, confirming that the structural similarity term provides a useful photometric signal beyond per-pixel L1.

Disabling depth supervision increases ATE from 1.77\,cm to 2.30\,cm, demonstrating that depth is critical for accurate pose estimation, though rendering quality remains largely unaffected. Removing the normal consistency loss has minimal impact on rendering metrics but slightly increases ATE, as it primarily aids geometric coherence rather than appearance. Removing the equilateral regulariser improves rendering metrics, at a modest cost to tracking accuracy. The equilateral loss discourages triangles from becoming slivers which affects tracking, however elongated triangles can fit better to irregularly shaped objects, thereby increasing visual fidelity. We include all hyperparameters used for the experiments in \cref{tab:hyperparams} and \cref{tab:hyperparams_tum}.

\begin{table}[h]
\centering
\setlength{\tabcolsep}{7pt}
\caption{Ablation study on \textbf{TUM fr1/desk}. We ablate each mapping loss component independently.}
\label{tab:ablation}
\begin{tabular}{lcccc}
\toprule
\textbf{Method} & \textbf{ATE [cm]} $\downarrow$ & \textbf{PSNR} $\uparrow$ & \textbf{SSIM }$\uparrow$ &\textbf{ LPIPS} $\downarrow$ \\
\midrule
Ours                  & \textbf{1.77} & 18.87 & 0.641 & 0.321 \\
w/o equilateral       & 1.90          & \textbf{19.54} & \textbf{0.667} & \textbf{0.289} \\
w/o normal            & 1.82          & 19.32 & 0.641 & 0.310 \\
w/o SSIM              & 2.24          & 18.83 & 0.620 & 0.349 \\
w/o depth             & 2.30          & 19.05 & 0.644 & 0.312 \\
\bottomrule
\end{tabular}
\end{table}

\vspace{5mm}

\cref{tab:ablation-avg} shows the contribution of densification, blur-split sensitivity, and opacity parameters on TUM. Densification improves ATE and rendering quality with minimal overhead. Mean vertex opacity reduces the triangle count and thus improves the runtime with similar ATE and rendering quality. Pruning more non-opaque triangles ($\rho=0.4$ vs $0.3$) decreases the expressivity, therefore more triangles are spawned. Hyperparameters such as keyframe-translation and densification thresholds should be scaled to suit the scene size.

\begin{table}[h]
  \centering
  \caption{Ablation study averaged over all three TUM sequences.}
  \label{tab:ablation-avg}
  \resizebox{\columnwidth}{!}{%
  \begin{tabular}{lcccccc}
  \toprule
  \textbf{Config} & \textbf{ATE (cm)} $\downarrow$ & \textbf{PSNR} $\uparrow$ & \textbf{SSIM} $\uparrow$ & \textbf{LPIPS} $\downarrow$ & \textbf{Tri.} & \textbf{FPS} $\uparrow$ \\
  \midrule
  No densification               & 1.94 & 17.58 & 0.665 & 0.336 & \textbf{25k} & \underline{1.47} \\
  Prune threshold $\rho{=}0.4$   & 1.61 & 17.70 & 0.678 & \textbf{0.321} & 48k & 1.43 \\
  Blur-split $\theta{=}0.0001$   & \underline{1.54} & 17.63 & \underline{0.681} & 0.332 & 30k & \underline{1.47} \\
  Blur-split $\theta{=}0.00005$ (Ours) & 1.55 & \underline{17.75} & 0.667 & 0.329 & \underline{28k} & \textbf{1.48} \\
  \quad + min opacity            & \textbf{1.52} & \textbf{17.91} & \textbf{0.691} & \underline{0.324} & 35k & 1.33 \\
  \bottomrule
  \end{tabular}%
  }
\end{table}

\begin{table}[t]
\centering
\caption{Hyperparameters used for the Replica dataset.}
\label{tab:hyperparams}
\setlength{\tabcolsep}{8pt}
\begin{tabular}{llc}
\toprule
\textbf{Category} & \textbf{Parameter} & \textbf{Value} \\
\midrule
\multirow{5}{*}{Tracking}  
 & Tracking iterations           & 100 \\
 & Rotation learning rate        & 0.003 \\
 & Translation learning rate     & 0.001 \\
 & D-SSIM weight $\lambda$       & 0.2 \\
 & Depth weight $\lambda_{dep}$ (tracking) & 0.05 \\
\midrule
\multirow{6}{*}{Mapping}
 & Iterations per view           & 30 \\
 & Init mapping iterations       & 150 \\
 & Pose learning rate scale      & 0.5 \\
 & Feature learning rate         & 0.0005 \\
 & Vertex learning rate (init)   & 0.0005 \\
\midrule
\multirow{4}{*}{Loss weights}
 & Equilateral $\lambda_{equi}$          & 1.2 \\
 & Normal $\lambda_{norm}$               & 0.05 \\
 & Normal (supervised) $\lambda_{norm,s}$ & 0.15 \\
 & Depth $\lambda_{dep}$ (mapping)        & 0.05 \\
\midrule
\multirow{4}{*}{Keyframing}
 & Keyframe interval             & 5 \\
 & Translation threshold         & 0.08\,m \\
 & Min translation               & 0.05\,m \\
 & Overlap threshold             & 0.95 \\
\midrule
\multirow{5}{*}{Densification}
 & Blur-split $\theta $          & 0.00005 \\
 & Gradient threshold            & 0.001 \\
 & Gradient percentile           & 0.99 \\
 & Max aspect ratio              & 5.0 \\
 & Min pixel count               & 4 \\
\midrule
\multirow{2}{*}{Pruning}
 & Opacity threshold $\epsilon_o$ & 0.3 \\
 & Max projected area $\epsilon_a$ & 400\,px \\
\midrule
\multirow{1}{*}{Rendering}
 & $\sigma$                      & 0.5 \\
\bottomrule
\end{tabular}
\end{table}

\begin{table}[h]
\centering
\caption{Hyperparameters used for the TUM RGB-D dataset. Other parameters remain the same as Replica.}
\label{tab:hyperparams_tum}
\setlength{\tabcolsep}{8pt}
\begin{tabular}{llc}
\toprule
\textbf{Category} & \textbf{Parameter} & \textbf{Value} \\
\midrule
\multirow{6}{*}{Mapping}
 & Iterations per view           & 30  \\
 & Init mapping iterations       & 250 \\
 & Pose learning rate scale      & 0.5 \\
 & Feature learning rate         & 0.00025 \\
 & Vertex learning rate (init)   & 0.0005 \\
\midrule
\end{tabular}
\end{table}

\end{document}